\pdfoutput=1
\documentclass[11pt]{article}
\usepackage{acl}
\usepackage[T1]{fontenc}
\usepackage[utf8]{inputenc}
\usepackage{times}
\usepackage{latexsym}
\usepackage{microtype}
\usepackage{inconsolata}
\usepackage{amsmath,amssymb,bm}
\usepackage{booktabs,multirow,makecell}
\usepackage{graphicx}
\usepackage{tabularx}
\usepackage[table]{xcolor}
\usepackage{xparse}
\usepackage{threeparttable}
\usepackage{fontawesome5}
\usepackage{array}
\usepackage{placeins}
\usepackage{url} 
\usepackage{natbib}
\usepackage{float}
\usepackage{graphicx}
\usepackage{pdfpages}

\usepackage{hyperref} 

\definecolor{ProprietaryColor}{RGB}{255,235,225}
\definecolor{ExcellentColor}{RGB}{220,245,220} 
\definecolor{GoodColor}{RGB}{235,245,255}      
\definecolor{FairColor}{RGB}{255,245,220}      
\definecolor{PoorColor}{RGB}{255,230,230}      

\definecolor{ClosedColor}{HTML}{D8EEF2}
\definecolor{OpenColor}{HTML}{FDEBDD}
\definecolor{HeaderColor}{RGB}{50,90,130}

\newcommand{\name}{\texttt{InfTool}}
\usepackage{amssymb}

\title{Close the Loop: Synthesizing Infinite Tool-Use Data via Multi-Agent Role-Playing}

\author{
  {\bf Yuwen Li}\textsuperscript{\rm 1},
  {\bf Wei Zhang}\textsuperscript{\rm 2 \thanks{\ Equal contribution. }},
  {\bf Zelong Huang}\textsuperscript{\rm 1},
  {\bf Mason Yang}\textsuperscript{\rm 1},
  {\bf Jiajun Wu}\textsuperscript{\rm 2},
  {\bf Shawn Guo}\textsuperscript{\rm 3},\\
  {\bf Huahao Hu}\textsuperscript{\rm 1},
  {\bf Lingyi Sun}\textsuperscript{\rm 1},
  {\bf Jian Yang}\textsuperscript{\rm 2†},
  {\bf Mingjie Tang}\textsuperscript{\rm 1}\thanks{\ Corresponding author.},
  {\bf Byran Dai}\textsuperscript{\rm 3},\\
   \textsuperscript{\rm 1}Sichuan University;  
   \textsuperscript{\rm 2}Beihang University;
   \textsuperscript{\rm 3}IQuest Research; \\
   \texttt{\{ywli\}@stu.scu.edu.cn;} \\
}

\begin{document}
\maketitle




\begin{abstract}

Enabling Large Language Models (LLMs) to reliably invoke external tools remains a critical bottleneck for autonomous agents. Existing approaches suffer from three fundamental challenges: \textit{expensive human annotation for high-quality trajectories, poor generalization to unseen tools, and quality ceilings inherent in single-model synthesis that perpetuate biases and coverage gaps}. We introduce \textbf{\name}, a fully autonomous framework that breaks these barriers through self-evolving multi-agent synthesis. Given only raw API specifications, \name~orchestrates three collaborative agents (User Simulator, Tool-Calling Assistant, and MCP Server) to generate diverse, verified trajectories spanning single-turn calls to complex multi-step workflows. The framework establishes a closed loop: synthesized data trains the model via Group Relative Policy Optimization (GRPO) with gated rewards, the improved model generates higher-quality data targeting capability gaps, and this cycle iterates without human intervention. Experiments on the Berkeley Function-Calling Leaderboard (BFCL) demonstrate that \name~transforms a base 32B model from 19.8\% to \textbf{70.9\% accuracy (+258\%)}, surpassing models 10× larger and rivaling Claude-Opus, and entirely from synthetic data without human annotation.

\end{abstract}

\section{Introduction}
\begin{figure}[t]
    \centering
    \includegraphics[width=0.8\columnwidth]{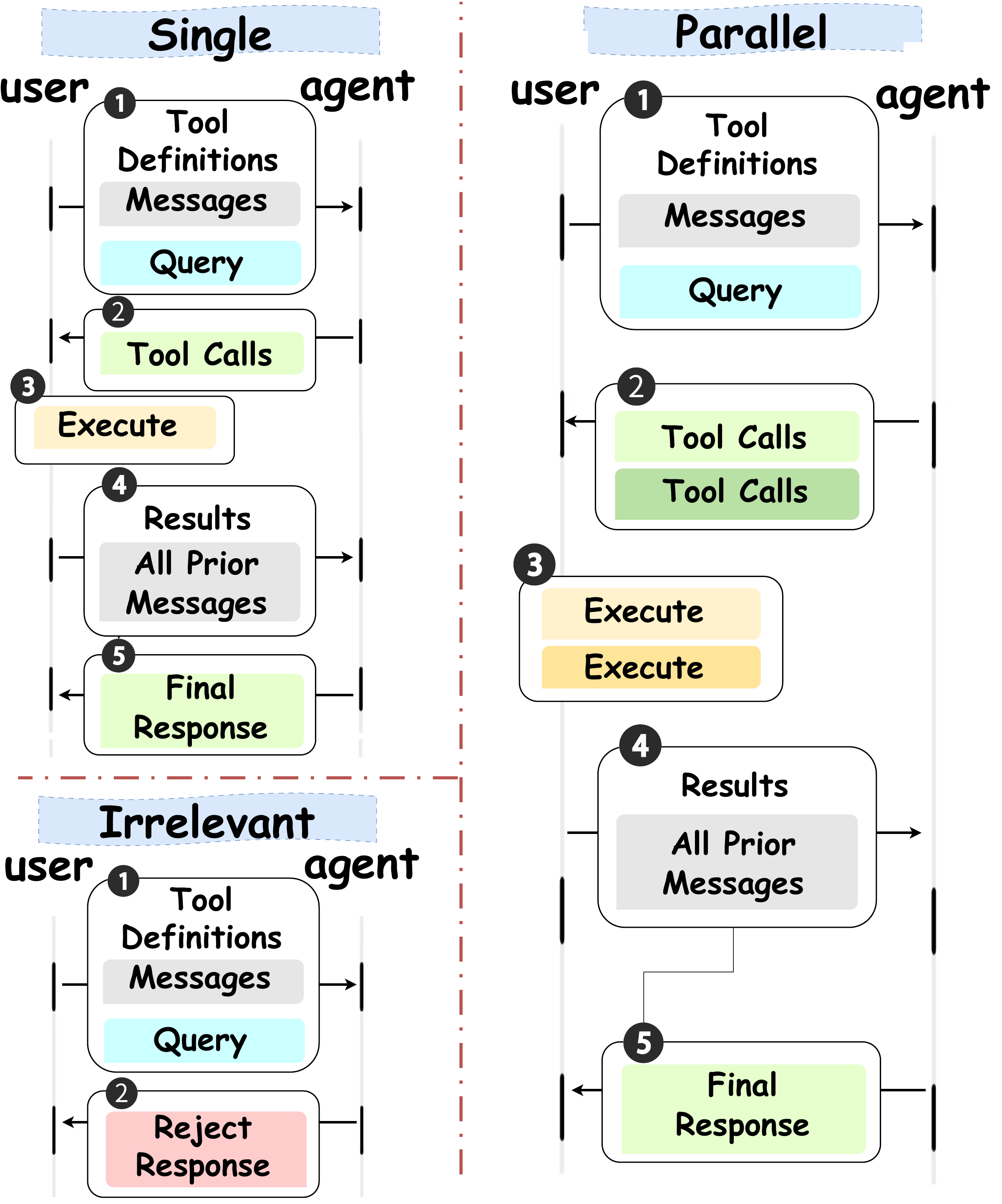}
    \vspace{-5pt}
    \caption{MCP-based tool calling across three scenarios.}
    \label{fig:intro}
    \vspace{-20pt}
\end{figure}

Tool calling~\cite{toolformer} in Figure~\ref{fig:intro} is pivotal for transforming LLMs into autonomous agents to interactive with external system and accomplish tasks beyond the intrinsic knowledge. However, reliable and generalizable tool callings remain 3 fundamental challenges: \textbf{1) Data Scarcity}, where high-quality interaction trajectories require expensive human annotation and post verification; \textbf{2) Limit Generalization}: Static datasets fail to adapt unseen tools during training; \textbf{3) Quality Cell}: synthesizing training data with LLMs  traps performance in a local optimum, inheriting synthesis model biases and blind coverage.

\begin{figure*}[t]
    \centering
    \includegraphics[width=0.95\textwidth]{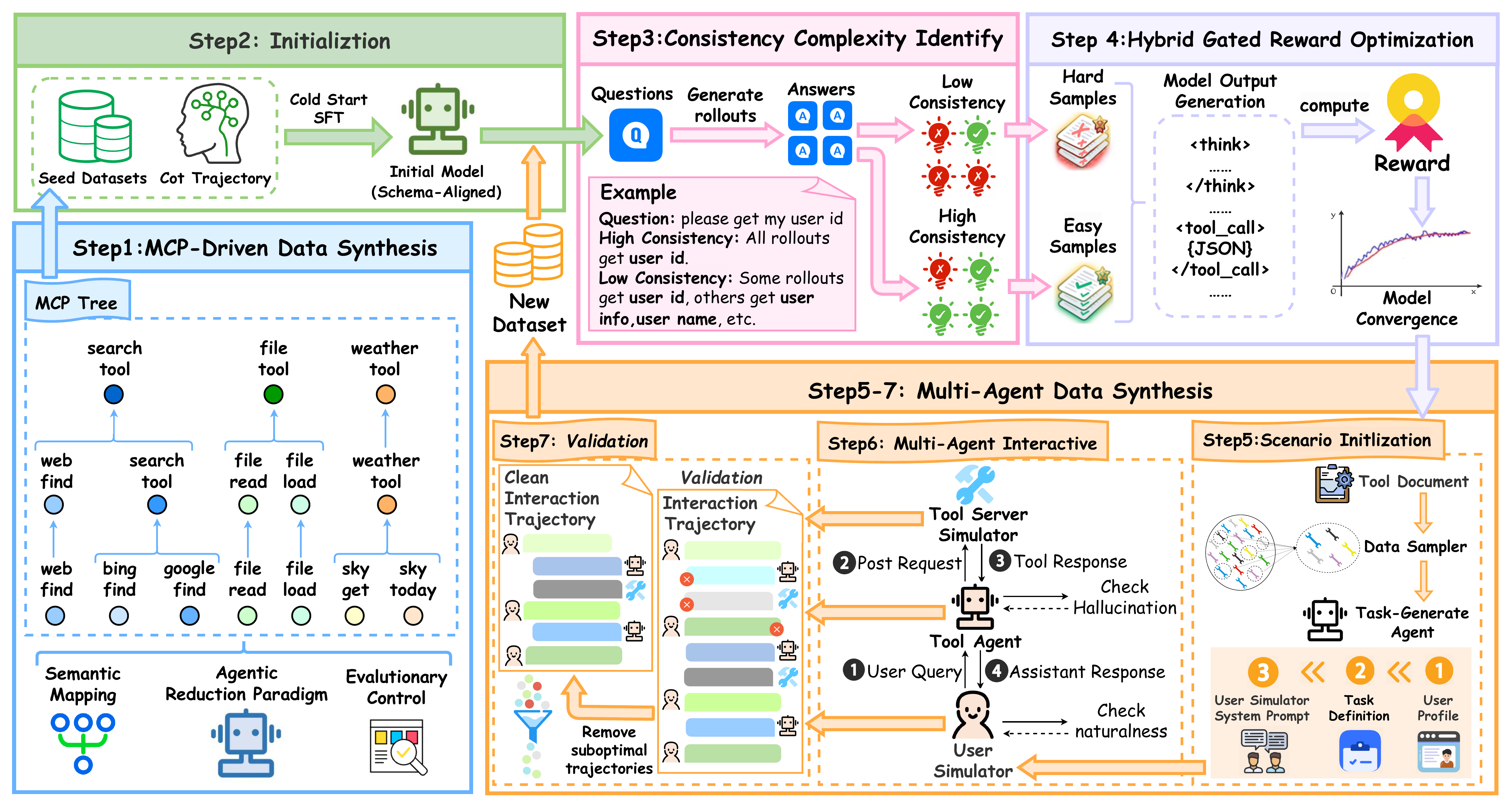}
    \vspace{-5pt}
    \caption{Overview of our framework. Step 1: Build MCP tool definitions from raw APIs. Step 2: Cold start Supervised Fine Tuning (SFT)~\cite{sft}. Step 3: Rollout evaluation to categorize samples. Step 4: GRPO training. Step 5-6: Multi-agent role-play (User, Agent, Server) generates complex trajectories. Step 7: Self-reflection filtering.}
    \label{fig:overall}
    \vspace{-20pt}
\end{figure*}

To address these challenges, we propose \textbf{\name}, a fully \textbf{Human-Annotation-Free} self-evolving framework that establishes a closed-loop system of synthetic data generation and iterative model optimization. Specifically, \name{} tackles \textbf{Data Scarcity} by employing a role-playing multi-agent system (User Simulator, Tool Agent, and Server Simulator) to automatically generate high-quality interaction  trajectories from raw API specifications, eliminating expensive human annotation. To overcome \textbf{Limited Generalization}, our framework synthesizes training data directly from MCP tool schemas, enabling adaptation to unseen tools without requiring static datasets. To break through the \textbf{Quality Ceiling}, we integrate GRPO~\cite{grpo} with a gated composite reward mechanism that ensures synthesized data quality continuously improves: the optimized model generates increasingly sophisticated trajectories, which in turn drives further policy refinement, creating a self-reinforcing virtuous cycle that transcends the limitations of single-model synthesis.

Our main contributions are as follows: \textbf{(1) Framework Innovation:} We introduce \name, a self-evolving framework that utilizes simulated multi-agent role-playing to generate high-quality tool-calling data purely from API specifications, eliminating the need for human annotation and post-hoc verification. \textbf{(2) Scalable Synthesis}: We propose a methodology for synthesizing training data at unbounded scale with controllable complexity through iterative GRPO-driven optimization, effectively resolving data scarcity and quality ceiling issues in multi-turn scenarios. \textbf{(3) State-of-the-Art Performance:} Experiments demonstrate  that \name{} improves the base model (Qwen2.5-32B) performance from 19.8 to 70.9 on BFCL (+\textbf{258.08\%}). Notably, \name{}-32B outperforms strong open-source baselines~\citep{deepseekv3,Qwen3coder} and achieves competitive performance with leading proprietary models~\citep{claude3,gpt4}.    


\section{Methodology}
\label{sec:methodology}

We first introduce the preliminary concepts of tool use and the RL~\citep{RL} which underpin our system. Subsequently, we present the \textsc{Role-Playing Multi-Agent Framework}~\citep{camel}, an integrated system designed for high-fidelity data synthesis. Finally, we elaborate on the iterative self-evolving mechanism that acts as the framework's engine, driving continuous capability enhancement.

\subsection{Preliminaries}
\label{sec:preliminaries}

\paragraph{Tool Use and MCP} Agentic tool use capability is formally defined as the policy $\pi$ mapping a user query $q$ and a context $C$ to a structured action space $\mathcal{A}$. In our work, we adopt the MCP, a standardized interface that decouples the AI model as the client from the data sources as the server. Within MCP, a tool call is represented as a tuple $t = (f, \mathbf{a})$, where $f$ is the function identifier and $\mathbf{a}$ represents the arguments conforming to a strict JSON schema $\mathcal{S}_f$. The fundamental challenge in MCP training is ensuring the model's generated $\hat{\mathbf{a}}$ strictly adheres to $\mathcal{S}_f$ while satisfying the semantic intent of $q$.

\paragraph{Group Relative Policy Optimization (GRPO)} To optimize the model's reasoning and tool-invocation precision, we employ GRPO~\citep{grpo}, a policy optimization algorithm that eliminates the need for a separate value network, thereby reducing memory overhead. GRPO optimizes the policy $\pi_\theta$ by maximizing the objective:\begin{equation}\begin{aligned}
\mathcal{J}(\theta) = \mathbb{E} \bigg[ \frac{1}{G} \sum_{i=1}^G \min \big( \rho_i A_i, \quad & \\
\hspace{-1.5em} \text{clip}(\rho_i, 1-\epsilon, 1+\epsilon) A_i \big) & - \beta \mathbb{D}_{\text{KL}}(\pi_\theta || \pi_{\text{ref}}) \bigg]\end{aligned}\end{equation}
where $\mathbb{E}$ denotes $\mathbb{E}_{q \sim P(Q), \{o_i\} \sim \pi_{\theta_{\text{old}}}}$, and $\rho_i = \frac{\pi_\theta(o_i|q)}{\pi_{\theta_{\text{old}}}(o_i|q)}$ is the importance ratio. Here, $\pi_{\theta_{\text{old}}}$ is the sampling policy, $G$ is the group size, and $A_i$ is the advantage. This compact formulation supports efficient iterative updates on self-generated data.

\subsection{Role-Playing Multi-Agent Data Synthesis}
\label{sec:multi_agent_framework}

\paragraph{Framework Overview} We propose a self-evolving role-playing multi-agent framework that unifies high-quality data synthesis with iterative model refinement. In Figure~\ref{fig:overall}, the framework operates as a closed loop:
1) \textbf{MCP Driven Data Synthesis} employs a multi-agent system to construct complex interaction trajectories from raw API definitions.
2) \textbf{Iterative Self-Evolving Training Loop} utilizes the synthesized data and GRPO to update the model, and the model in turn empowers the agents to generate even higher-quality data in subsequent iterations.

\paragraph{MCP Tree}
\label{sec:synthesis_details}
We initialize the pipeline by ingesting 17,713 candidates from RapidAPI\footnote{\url{https://rapidapi.com/hub}}. We propose an iterative refinement framework for a rigorous foundation, designed to disambiguate unclear MCP Tool definitions and merge functionally redundant entries. Formally, let $\mathcal{T}^{(r)}$ denote the toolset at the $r$-th iteration. For each tool $t_i \in \mathcal{T}^{(r)}$, we compute its semantic embedding $\mathbf{e}_i$. Tools exhibiting high semantic overlap are aggregated into $K$ candidate clusters $\mathcal{C} = \{\mathcal{C}_1, \dots, \mathcal{C}_K\}$. A cluster $\mathcal{C}_k$ is defined as a subset where elements maintain a pairwise cosine similarity exceeding a threshold $\delta$:
\begin{equation}
\small 
    \mathcal{C}_k = \{ t_i \in \mathcal{T}^{(r)} \mid \exists t_j \in \mathcal{C}_k, j \neq i, \cos(\mathbf{e}_i, \mathbf{e}_j) > \delta \}
\end{equation}Subsequently, an LLM acts as a semantic discriminator to partition $\mathcal{C}_k$ into a redundant subset $\mathcal{R}_k$ and a unique subset $\mathcal{U}_k$. The refined toolset $\mathcal{T}^{(r+1)}$ is derived by synthesizing a consolidated definition $\hat{t}$ for the redundant subset via a generative abstraction operator $\mathcal{A}_{\text{gen}}$:
\begin{equation}
    \mathcal{T}^{(r+1)} = \bigcup_{k=1}^{K} \left( \mathcal{U}_k \cup \{ \mathcal{A}_{\text{gen}}(\mathcal{R}_k) \} \right)
\end{equation}
This procedure yields 3,059 high-fidelity synthetic MCP Tools.

\paragraph{Hierarchical Data Synthesis} We synthesize the data in two stage.  
\textbf{Single-Turn Synthesis}
Single-turn synthesis addresses three specific sub-tasks: standard execution, parallel tool execution, and irrelevance detection. We implement a streamlined pipeline where selected tool schemas serve as context for generating user queries. An agent executes the task, producing a trajectory subjected to rigorous verification.
\textbf{Multi-Turn Synthesis via Role-Playing}
For complex scenarios, we developed an interactive multi-agent playground populated by three roles: an User Simulator, a Tool Agent (MCP Client), and a MCP Server Simulator.
A Task Generation Agent creates dynamic scenarios defined by five dimensions: user profile, known/unknown information, user need, and difficulty. The agents interact to resolve unknown information progressively, generating 3 to 5 round interaction trajectories that enforce strict logical dependencies.

\paragraph{Quality Assurance Mechanism} We implement a dual-layer quality assurance mechanism to guarantee fidelity: \textbf{1) Intrinsic Self-Reflection (In-Trajectoral):}
Operating in real-time, this stage enforces stylistic constraints $\Psi_{\text{style}}$ for the user agent and strict MCP schema compliance for the client agent. The corrected response $o'_t$ is generated via a self-correction function 
\begin{equation}
    o'_t = \mathcal{S}(o_t \mid \Psi_{\text{style}}(o_t), \Psi_{\text{schema}}(o_t))
\end{equation}
This effectively rectifies errors such as hallucinated tool requests or argument violations. \textbf{2) Multi-Agent Validation (Post-Trajectoral):}
Upon trajectory completion, validator agents $V = \{v_1, \dots, v_M\}$ leverage a voting-deliberation-correction loop. The final accepted trajectory $\mathcal{L}^*$ is derived by minimizing redundancy $R(\cdot)$ through consensus:
\begin{equation}
    \mathcal{L}^* = \mathcal{M}_{\text{consensus}}(\mathcal{L} \mid \{v_i\}_{i=1}^{M})
\end{equation}

\begin{figure*}[ht!]  
\centering
\includegraphics[width=0.95\textwidth]{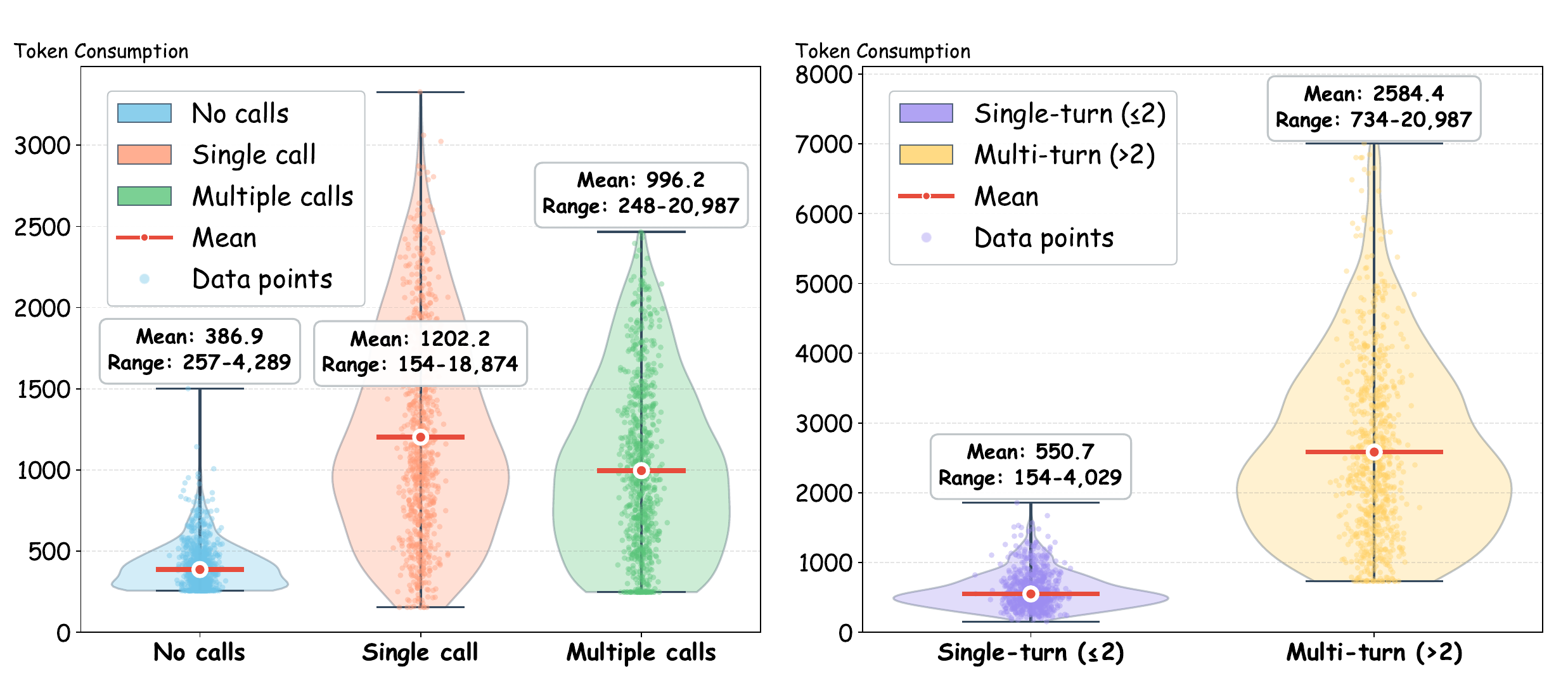}
\vspace{-10pt}
\caption{Token consumption distribution across tool-call categories and conversation lengths.  The left panel shows token requirements for different tool-call frequencies (no calls, single call, multiple calls),  while the right panel contrasts single-turn versus multi-turn conversations. Box plots indicate quartiles, with whiskers extending to the 5th and 95th percentiles.}
\label{fig:data-distribution}
\vspace{-15pt} 
\end{figure*}

\subsection{Self-Evolution Training Loop}
\label{sec:training_loop}

To enable the framework to progressively enhance its own capabilities, we integrate a self-evolving training loop. This component is not merely a downstream application but the engine that powers the ``Self-Evolving'' nature of our role-playing framework.

\paragraph{Complexity Stratification} At each iteration $t$, the agent's policy $\pi_{\theta_t}$ is evaluated to stratify the dataset $\mathcal{D}_t$ into ``Easy'' and ``Hard'' subsets based on an execution consistency score $c_i$:
\begin{equation}
\mathcal{D}_t^{\text{Hard}} = \{ (\mathbf{q}_i, \mathbf{a}_i^*) \mid c_i < \gamma \}
\end{equation}
The ``Hard'' subset defines the learning frontier for the next cycle.

\paragraph{Gated Composite Reward Policy Optimization} We update the policy using GRPO with a specialized composite reward function $R(\mathbf{T})$ that aligns the model with the MCP protocol:
\begin{equation}
\begin{aligned}
R(\mathbf{T}) =\;& R_{\text{format}}(\mathbf{T}) 
+ R_{\text{tool}}(\mathbf{a}, \mathbf{a}^*) \\
&+ \mathbb{I}\big(R_{\text{tool}} > \tau_t\big)\cdot \alpha \cdot R_{\text{teacher}}(\mathbf{r}; \mathbf{q})
\end{aligned}
\end{equation}here, $R_{\text{format}}$ rewards syntactic validity (e.g., correct \texttt{<think>} and \texttt{<tool\_call>} tags), while $R_{\text{tool}}$ measures execution fidelity. Crucially, a reasoning quality reward $R_{\text{teacher}}$ is \textit{gated} ($\mathbb{I}$), meaning it is only awarded when correct execution, strictly binding reasoning improvements to functional utility.

\paragraph{Adaptive Data Injection} The cycle closes by engaging the \textit{Multi-Agent Data Synthesis Module} (described in Sec.~\ref{sec:synthesis_details}) to generate new training samples $\mathcal{D}_t^{\text{new}}$ specifically targeting the identified competency frontiers in $\mathcal{D}_t^{\text{Hard}}$. This creates a self-sustaining loop where improved agents generate superior data, which in turn trains stronger agents.

\begin{figure}[t]
    \centering
    \includegraphics[width=0.9\linewidth]{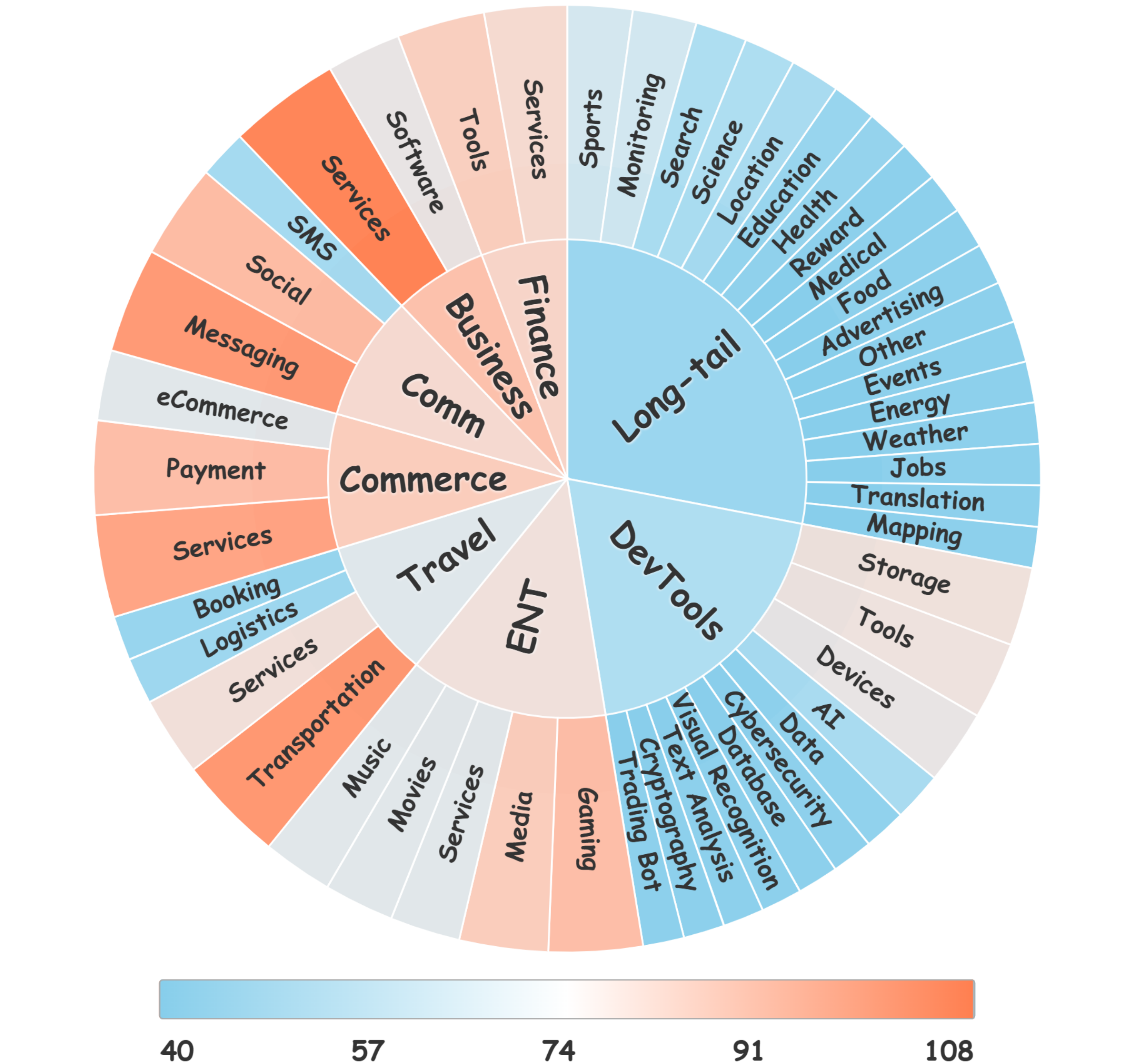}
    \caption{Domain distribution in the training dataset.}
    \label{fig:domain_sunburst}
    \vspace{-15pt}
\end{figure}

\begin{figure}[t]
    \centering
    \includegraphics[width=0.95\linewidth]{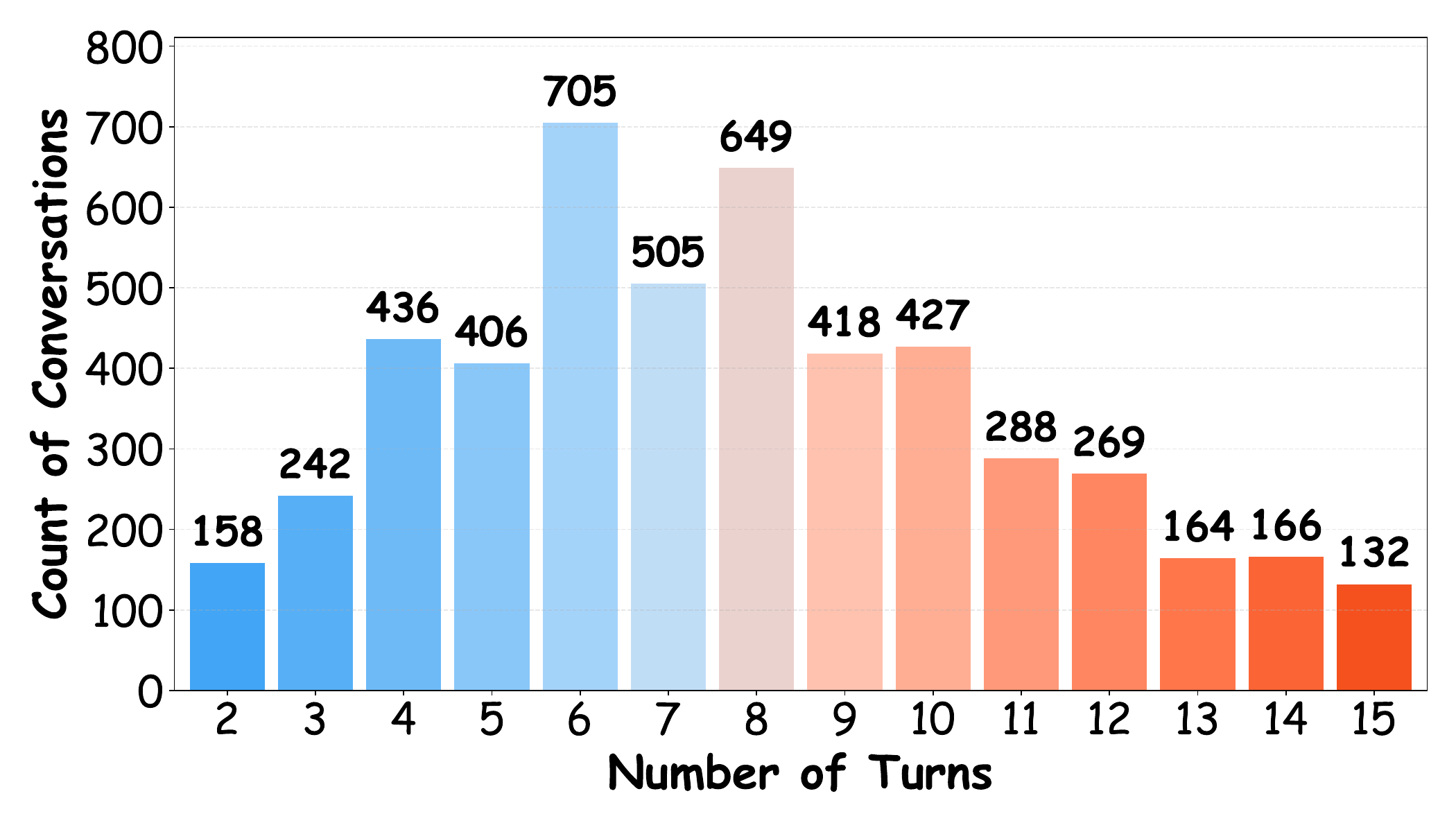}
    \vspace{-10pt}
    \caption{Distribution of Conversation Turns.}
    \label{fig:turn}
    \vspace{-20pt}
\end{figure}

\section{Data Analysis}

\paragraph{Computational Complexity Analysis} Analysis identifies three interaction patterns with distinct computational demands. No tool-calls are simplest, while single tool-calls, comprising 53.7\% of the dataset, are more demanding. Multi tool-calls show similar average complexity but greater consistency. Conversation length further amplifies load: multi-turn dialogues require approximately four times the tokens of single-turn ones in Figure~\ref{fig:data-distribution}.

\paragraph{Dynamic Dataset Composition and Evolution} Our training corpus evolves dynamically through the self-evolving RL loop. The initial seed dataset spans 15 primary domains and a long-tail of 35 additional categories, with broad coverage across diverse application scenarios. Critical baseline analysis reveals that 82.83\% of MCP tools require parameter specifications, establishing a significant benchmark of complexity for semantic parsing. Domain-specific analysis indicates particularly high parameter requirements in structured fields such as Finance, Travel, and Sports in Figure~\ref{fig:domain_sunburst}.

\paragraph{Multi-Turn Conversation Statistics} We further analyze multi-turn dialogues (with more than one turn) in our dataset, totaling 4,965 conversations. The number of turns ranges from 2 to 15, with an average of 7.78 turns and a median of 8 turns. Most conversations are concentrated in the 6--8 turn range, which accounts for 37.44\% of the dataset. Overall, 62.46\% of multi-turn dialogues have 8 turns or fewer, while longer conversations exceeding 10 turns represent 20.52\%, highlighting the presence of substantial long-horizon interactions that challenge contextual maintenance and sequential tool usage in Figure~\ref{fig:turn}.

\begin{table*}[t]
  \centering
  \small
  \renewcommand{\arraystretch}{1.3} 
  \vspace{-10pt}
  \resizebox{\textwidth}{!}{%
    \begin{tabular}{l r c | cccc | cccc | cccc | c c}
      \toprule
      
      \multirow{3}{*}{\textbf{Model}} & 
      \multirow{3}{*}{\textbf{Size}} & 
      \multirow{3}{*}{\textbf{Total}} & 
      \multicolumn{4}{c|}{\textbf{Multi-Turn}} & 
      \multicolumn{4}{c|}{\textbf{Non-Live}} & 
      \multicolumn{4}{c|}{\textbf{Live}} & 
      \multirow{3}{*}{\textbf{Relevance}} & 
      \multirow{3}{*}{\textbf{Irrelevance}} \\
      
      \cmidrule(lr){4-7} \cmidrule(lr){8-11} \cmidrule(lr){12-15}
      
      & & & 
      Base & \makecell{Miss\\Func} & \makecell{Miss\\Param} & \makecell{Long\\Cont.} & 
      Simple & Multiple & Parallel & \makecell{Multi\\Para.} & 
      Simple & Multiple & Parallel & \makecell{Multi\\Para.} & & \\
      
      \midrule

      \rowcolor{HeaderColor}
      \multicolumn{17}{c}{\textcolor{white}{\textbf{Closed-source Model}}} \\
      \midrule

      \rowcolor{ExcellentColor}
      \small Claude-Opus-4-5-20251101 & \faLock{} & \textbf{78.9} &
      \textbf{81.0} & \underline{64.0} & \textbf{58.0} & \textbf{70.5} &
      \underline{76.8} & \textbf{95.5} & \underline{93.5} & 88.5 &
      86.5 & 78.1 & \textbf{87.5} & 75.0 &
      62.5 & 84.7 \\

      \rowcolor{ExcellentColor}
      \small Claude-Sonnet-4-5-20250929 & \faLock{} & \underline{77.0} &
      69.0 & \textbf{65.0} & 52.5 & 59.0 &
      72.6 & \textbf{95.5} & \textbf{94.5} & \textbf{92.0} &
      \textbf{89.5} & \underline{78.9} & \textbf{87.5} & \textbf{83.3} &
      68.7 & \textbf{86.6} \\

      \rowcolor{ExcellentColor}
      \small Gemini-3-Pro-Preview & \faLock{} & 76.8 &
      69.0 & 63.0 & \underline{56.5} & \underline{64.0} &
      75.5 & \underline{94.0} & 91.0 & 82.5 &
      \underline{87.6} & \textbf{80.4} & \underline{75.0} & \underline{79.1} &
      \underline{75.0} & 77.8 \\

    \rowcolor{ExcellentColor}
      \small GPT-5.2-2025-12-11 & \faLock{} & 60.4 &
      36.5 & 18.0 & 27.5 & 30.5 &
      72.9 & 88.0 & 89.0 & 77.5 &
      71.7 & 70.3 & 68.7 & 58.3 &
      \underline{75.0} & 79.4 \\

      \rowcolor{ExcellentColor}
      \small GPT-o3-2025-04-16 & \faLock{} & 40.4 &
      16.5 & 11.5 & 14.5 & 16.5 &
      74.5 & 87.0 & 0.0 & 0.0 &
      70.5 & 67.6 & 0.0 & 0.0 &
      \textbf{81.2} & \underline{86.1} \\

      \rowcolor{ExcellentColor}
      \small Grok-4-1-fast-reasoning & \faLock{} & 75.2 &
      \underline{70.5} & 59.5 & 43.0 & 62.5 &
      \textbf{77.5} & 93.0 & 92.5 & \underline{90.0} &
      84.1 & 77.3 & \underline{75.0} & 75.0 &
      \textbf{81.2} & 79.4 \\

      \midrule
      \rowcolor{HeaderColor}
      \multicolumn{17}{c}{\textcolor{white}{\textbf{Open-source 14B+ Model}}} \\
      \midrule
      \rowcolor{GoodColor}
      \small Kimi-K2-Instruct & 32B/1TB & \textbf{70.3} &
      \textbf{62.0} & \underline{41.0} & \textbf{44.5} & \textbf{55.0} &
      69.4 & 92.0 & 82.0 & 83.0 &
      81.7 & 78.0 & \underline{87.5} & 66.6 &
      75.0 & \underline{87.3} \\
      
      \rowcolor{GoodColor}
      \small DeepSeek-V3.2-exp & 685B & 41.96 &
      41.5 & 39.5 & \underline{33.5} & 35.0 &
      37.9 & 74.0 & 15.0 & 12.5 &
      66.2 & 51.6 & 25.0 & 25.0 &
      37.5 & \textbf{93.1} \\

      \rowcolor{GoodColor}
      \small Qwen3-Coder-480B-A35B-Instruct & 35B/480B & \underline{69.9} &
      51.5 & 27.5 & 31.0 & 49.5 &
      \textbf{77.9} & 94.0 & \textbf{93.0} & 86.0 &
      83.7 & \textbf{81.0} & 81.3 & 70.8 &
      66.7 & 85.5 \\

      \rowcolor{GoodColor}
      \small Qwen3-235B-A22B-Instruct & 22B/235B & 50.5 &
      \underline{54.0} & \textbf{42.5} & 31.5 & \underline{50.5} &
      40.5 & 36.5 & 53.0 & 19.5 &
      58.5 & 71.6 & 68.7 & 62.5 &
      87.5 & 81.7 \\

      \rowcolor{GoodColor}
      \small Qwen2.5-72B-Instruct & 72B & 63.8 &
      28.0 & 34.0 & 27.5 & 29.0 &
      73.6 & \underline{95.0} & 92.0 & 88.5 &
      74.0 & 77.5 & 50.0 & 62.5 &
      72.2 & 80.5 \\


      \rowcolor{GoodColor}
      \small Llama-3.3-70B-Instruct & 70B & 62.0 &
      28.0 & 19.5 & 14.0 & 24.5 &
      76.8 & \underline{95.0} & 90.0 & \textbf{92.0} &
      81.0 & 75.3 & \underline{87.5} & 75.0 &
      \textbf{100.0} & 53.7 \\

      \rowcolor{GoodColor}
      \small CoALM-70B & 70B & 65.1 &
      11.0 & 14.0 & 9.0 & 8.5 &
      70.2 & 92.0 & 88.5 & 83.0 &
      70.5 & 66.5 & 68.7 & 62.5 &
      \underline{93.7} & 85.6 \\

      \rowcolor{GoodColor}
      \small mistral-large-2411  & 70B & 60.2 &
      18.5 & 11.5 & 13.0 & 13.5 &
      72.0 & 93.5 & 89.5 & 83.5 &
      \textbf{87.2} & \underline{80.7} & 81.2 & \underline{75.0} &
      \underline{93.7} & 68.9 \\

      \rowcolor{GoodColor}
      \small Qwen3-32B Chat & 32B & 60.0 &
      19.5 & 16.5 & 17.0 & 13.5 &
      76.7 & \textbf{96.0} & 89.5 & \underline{91.5} &
      \underline{86.1} & \textbf{81.0} & 62.5 & 70.8 &
      87.5 & 75.7 \\

      \rowcolor{GoodColor}
      \small Qwen2.5-32B-Instruct & 32B & 62.2 &
      32.0 & 25.5 & 24.0 & 17.0 &
      65.5 & 92.5 & 92.0 & 89.0 &
      78.2 & 77.9 & 50.0 & 58.3 &
      66.6 & 80.2 \\

      \rowcolor{GoodColor}
      \small Qwen3-Coder-30B-A3B-Instruct & 3B/30.5B & 62.6 &
      36.0 & 16.0 & 19.5 & 33.0 &
      75.3 & \underline{95.0} & 88.0 & 83.5 &
      83.3 & 79.5 & 62.5 & \textbf{83.3} &
      88.9 & 77.0 \\

      \rowcolor{GoodColor}
      \small Gemma-3-27b  & 27B & 57.4 &
      16.5 & 4.5 & 8.0 & 14.0 &
      \underline{77.6} & 92.5 & 89.0 & 89.5 &
      84.5 & 72.4 & \textbf{93.7} & 45.8 &
      81.2 & 73.6 \\

      \rowcolor{GoodColor}
      \small Granite-20b & 20B & 48.8 &
      9.0 & 3.0 & 6.5 & 3.0 &
      72.9 & 91.5 & 83.5 & 81.5 &
      67.8 & 56.7 & 43.7 & 58.3 &
      87.5 & 75.1 \\

      \rowcolor{GoodColor}
      \small Qwen2.5-14B-Instruct & 14B & 59.9 &
      29.5 & 25.5 & 21.5 & 19.5 &
      65.0 & \textbf{96.0} & 79.5 & 86.5 &
      75.2 & 74.2 & 68.8 & 62.5 &
      72.2 & 76.4 \\

      \rowcolor{GoodColor}
      \small Qwen3-14B Chat & 14B & 58.5 &
      17.0 & 15.5 & 15.5 & 8.5 &
      74.4 & \underline{95.0} & \underline{92.5} & 88.0 &
      85.3 & 79.1 & 56.3 & 75.0 &
      \underline{75.0}75 & 82.9 \\

      \midrule
      \rowcolor{HeaderColor}
      \multicolumn{17}{c}{\textcolor{white}{\textbf{Open-source 0.5--14B Models}}} \\
      \midrule

      \rowcolor{FairColor}
      \small Bielik-11B-v2.3-Instruct & 11B & 50.6 &
      4.5 & 0.5 & 3.0 & 2.5 &
      \underline{73.0} & 92.0 & 85.5 & 75.5 &
      75.5 & 66.1 & 62.5 & 58.3 &
      \underline{93.7} & 36.0 \\

      \rowcolor{FairColor}
      \small Falcon3-10B-Instruct & 10B & 55.6 &
      6.5 & 9.5 & 5.0 & 5.0 &
      70.5 & \underline{93.5} & 88.5 & 87.5 &
      77.1 & 76.1 & 50.0 & 41.6 &
      \underline{93.7} & 32.0 \\

      \rowcolor{FairColor}
      \small Qwen3-8B Chat & 8B & \underline{59.5} &
      14.5 & 15.5 & \underline{13.5} & 12.0 &
      71.7 & \textbf{97.0} & 90.0 & \textbf{91.0} &
      \textbf{84.9} & \textbf{79.2} & \underline{68.8} & \underline{66.7} &
      81.3 & 78.1 \\

      \rowcolor{FairColor}
      \small Granite-3.1-8B-Instruct  & 8B & 48.7 &
      11.5 & 2.0 & 7.5 & 9.0 &
      67.3 & 92.0 & 84.0 & 70.0 &
      58.5 & 61.8 & 18.7 & 41.6 &
      68.7 & 79.9 \\

      \rowcolor{FairColor}
      \small CoALM-8B & 8B & 53.2 &
      10.0 & 7.0 & 8.0 & 7.0 &
      69.5 & \underline{93.5} & 88.0 & 88.5 &
      70.5 & 66.1 & 62.5 & 54.1 &
      87.5 & \underline{86.9} \\


      \rowcolor{FairColor}
      \small Hammer2.1-7b  & 7B & \textbf{59.6} &
      \underline{24.0} & \textbf{28.5} & \textbf{21.5} & \underline{21.5} &
      72.5 & 92.5 & \underline{91.0} & 86.0 &
      66.6 & 69.9 & \textbf{75.0} & \textbf{75.0} &
      50.0 & \textbf{90.1} \\

      \rowcolor{FairColor}
      \small Qwen3-4B Chat & 4B & 50.9 &
      9.5 & 9.0 & 7.0 & 9.5 &
      \textbf{75.2} & 93.0 & \textbf{92.5} & \underline{90.0} &
      \underline{78.3} & \underline{76.3} & 56.3 & \underline{66.7} &
      87.5 & 84.9 \\

      \rowcolor{FairColor}
      \small MiniCPM3-4B-FC & 4B & 51.2 &
      6.5 & 2.0 & 4.5 & 2.5 &
      70.5 & 92.0 & 84.0 & 80.5 &
      73.2 & 63.5 & 50.0 & 62.5 &
      68.7 & 72.8 \\

      \rowcolor{FairColor}
      \small Qwen2.5-3B-Instruct & 3B & 49.7 &
      9.0 & 0.0 & 8.0 & 3.5 &
      71.4 & 91.0 & 72.5 & 73.0 &
      71.3 & 70.0 & 62.5 & 54.2 &
      77.8 & 67.3 \\

      \rowcolor{FairColor}
      \small Llama-3.2-3B-Instruct & 3B & 48.3 &
      4.5 & 3.5 & 4.0 & 3.5 &
      70.7 & 92.0 & 88.5 & 77.0 &
      64.7 & 57.7 & 18.8 & 37.5 &
      \textbf{93.8} & 52.1 \\

      \rowcolor{FairColor}
      \small Falcon3-3B-Instruct & 3B & 36.7 &
      1.5 & 0.5 & 0.5 & 1.5 &
      56.5 & 69.5 & 67.0 & 25.5 &
      57.3 & 54.7 & 25.0 & 33.3 &
      81.2 & 32.9 \\

      \rowcolor{FairColor}
      \small Qwen2.5-1.5B-Instruct & 1.5B & 45.7 &
      3.5 & 1.0 & 2.5 & 2.0 &
      59.5 & 84.0 & 67.0 & 64.5 &
      69.4 & 59.5 & 50.0 & 62.5 &
      83.3 & 72.4 \\

      \rowcolor{FairColor}
      \small Arch-Agent-1.5B & 1.5B & 59.0 &
      \textbf{35.5} & \underline{27.5} & \textbf{21.5} & \textbf{22.0} &
      72.1 & 92.0 & 85.5 & 81.0 &
      70.5 & 67.8 & 31.2 & 58.3 &
      75.0 & 74.8 \\

      \rowcolor{FairColor}
      \small Llama-3.2-1B-Instruct & 1B & 16.7 &
      0.0 & 0.0 & 0.0 & 0.0 &
      44.1 & 50.0 & 43.5 & 14.0 &
      30.2 & 6.8 & 0.0 & 0.0 &
      43.8 & 53.0 \\

      \rowcolor{FairColor}
      \small Qwen3-0.6B Chat & 0.6B & 40.5 &
      2.5 & 1.5 & 0.0 & 1.0 &
      64.3 & 82.0 & 69.0 & 59.0 &
      58.1 & 50.7 & 56.3 & 41.7 &
      68.8 & 81.8 \\

      \rowcolor{FairColor}
      \small Qwen2.5-0.5B-Instruct & 0.5B & 35.7 &
      1.0 & 0.0 & 0.5 & 1.0 &
      48.4 & 74.5 & 48.5 & 43.0 &
      50.0 & 36.6 & 56.3 & 25.0 &
      72.2 & 64.6 \\

      \midrule
      \rowcolor{HeaderColor}
      \multicolumn{17}{c}{\textcolor{white}{\textbf{Baselines \& Our Models}}} \\
      \midrule
      \rowcolor{PoorColor}
      \small \name{}-32B & 32B & 70.9 &
      59.0 & 29.5 & 37.0 & 50.5 &
      70.3 & 95.0 & 93.5 & 91.5 &
      78.7 & 81.7 & 87.5 & 79.2 &
      66.6 & 88.9 \\

      \rowcolor{PoorColor}
      \small \name{}-7B & 7B & 61.7 &
      34.0 & 10.5 & 22.0 & 24.5 &
      59.5 & 89.5 & 89.0 & 90.0 &
      64.7 & 68.9 & 75.0 & 87.5 &
      44.4 & 93.9 \\



      \bottomrule
    \end{tabular}%
  }%
  \caption{Performance comparison on the Berkeley Function Calling Leaderboard (BFCL). ``Total'' denotes the overall accuracy.Following the leaderboard structure,metrics are categorized into Multi-Turn (including Base and error analysis).Non-Live (AST evluation),Live (Real Record), and Hallucination Measurement.Scores shown in boldface indicate the highest score for each metric.The \faLock{} icon indicates that the model's parameter size is not publicly released (size locked).``-'' indicates that the model was not explicitly trained for tool-calling capabilities and ``!'' indicates that the score is actually wrong (higher than real score).}
  \label{tab:new_results}
  \vspace{-15pt}
\end{table*}

\section{Experiment}
\subsection{Experiment Setup}
\label{sec:experimental_setup}
\paragraph{Hardware Infrastructure} We conduct our experiments on a robust computational infrastructure running Ubuntu 22.04, equipped with an Intel Xeon (R) Gold 6348 CPU @2.60GHz, 8 NVIDIA H800 GPUs, and 528 GB of memory, providing substantial computational resources for large scale model training and inference operations across multiple experimental phases.

\paragraph{Model Deployment} Our experimental pipeline employs three models for different phases: EmbeddingGemma for semantic clustering, Qwen3-Coder-30B-Instruct for tool refinement and data generation, and DeepSeek-R1 for chain-of-thought trajectory generation during cold-start initialization. We conduct supervised fine-tuning and iterative self-evolving RL across two model scales: Qwen2.5-7B and Qwen2.5-32B.

\paragraph{Training Setup} During SFT stage, we adopt a full parameter training strategy with 512 global batchsize, 1e-5 learning rate and train for 3 epochs under BF16~\citep{bf16} precision to improve both speed and numerical stability.
During the RL stage, we adopt a data sampling strategy with 0.7 generation temperature and generate 1000 steps per iteration at 1024 global batchsize.

\subsection{Evaluate Metric}
To rigorously assess multi-agent function calling capabilities, we evaluate our models on BFCL~\citep{berkeley}. BFCL V3 is chosen for its advanced multi-turn and multi-step assessment capabilities, which test iterative problem solving in complex scenarios like state inconsistencies. \textbf{Notes:} Although BFCL released version 4 earlier, technical issues prevented us from reproducing most of the results reported for this version in our experiments. Specifically, all newly added test items yielded zero values. Therefore, we restrict our experimental evaluation to version 3.

\begin{figure}[t]
    \centering
    \includegraphics[width=0.9\linewidth]{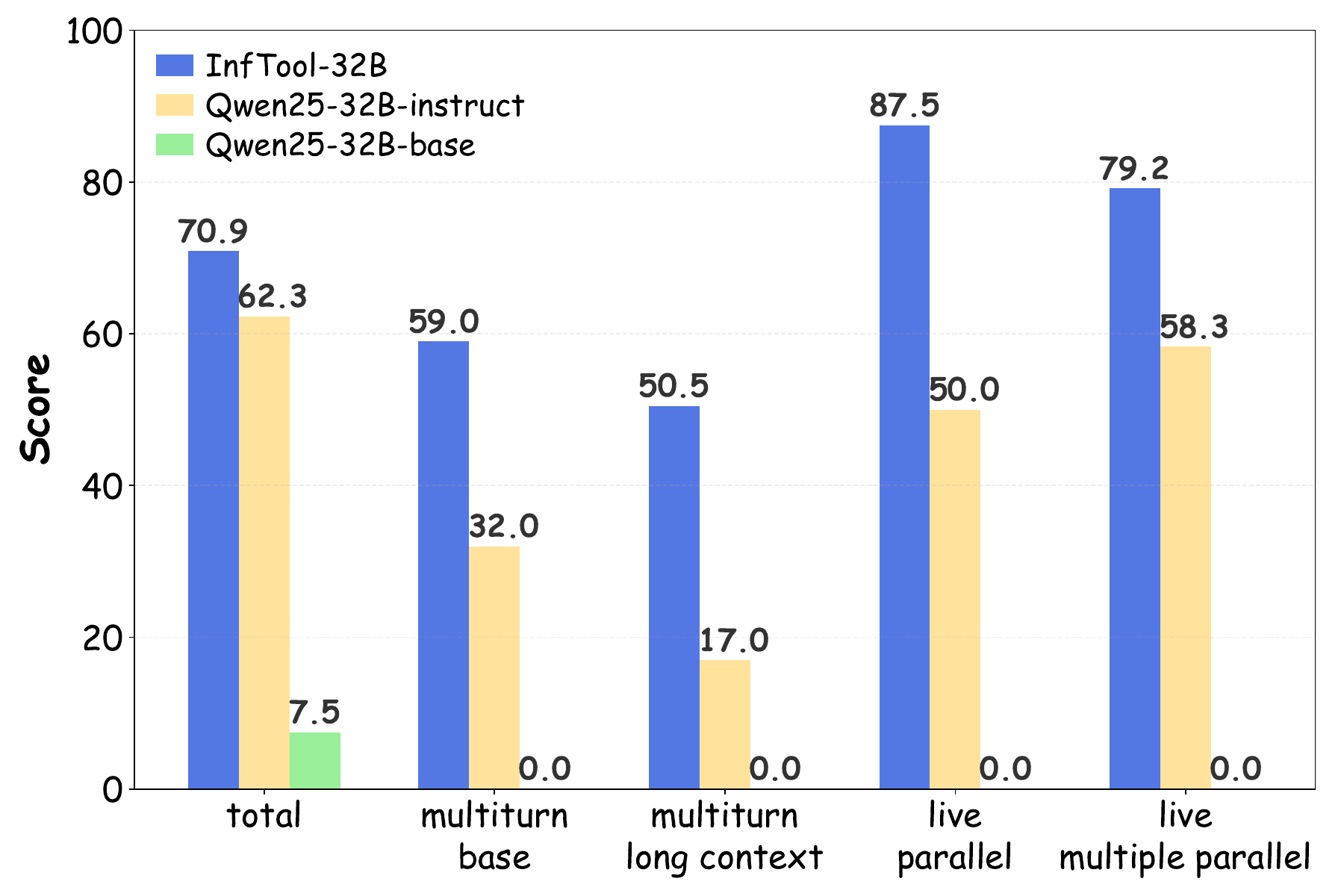}
    \vspace{-10pt}
    \caption{Comparison of \name{} and baselines.}
    \label{fig:compare}
    \vspace{-10pt}
\end{figure}

\begin{figure}[t]
    \centering
    \includegraphics[width=0.95\linewidth]{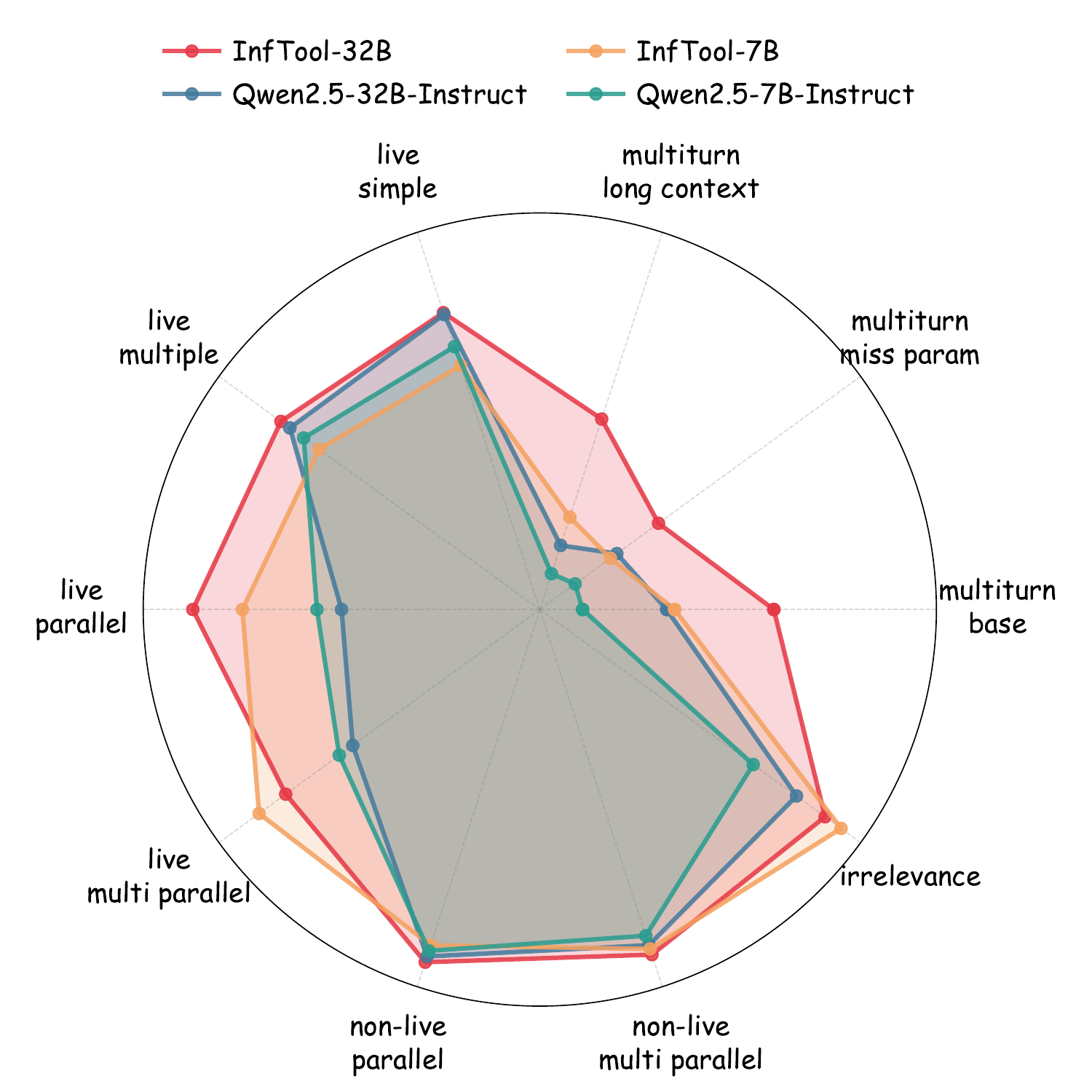} 
    \vspace{-10pt}
    \caption{Analysis of \name{} and baselines.}
    \label{fig:radar}
    \vspace{-20pt}
\end{figure}

\begin{figure*}[t]
    \centering
    \includegraphics[width=0.85\linewidth]{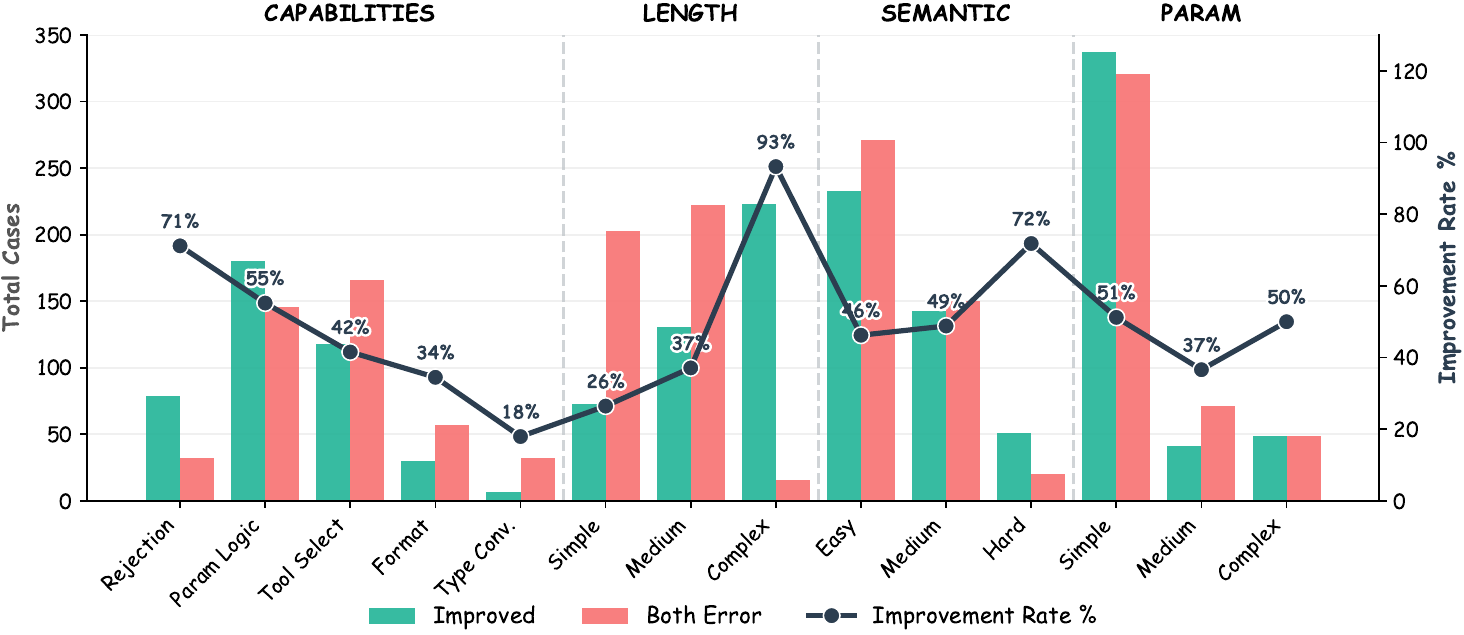} 
    \vspace{-5pt}
    \caption{\textbf{Unified Performance Landscape: Volume vs. Efficiency.} 
    This dual-axis chart integrates capability metrics with complexity dimensions. 
    The \textbf{Bars} represent the absolute volume of cases, divided into Improved samples (green) and Persistent Errors (red, indicating failures in both Instruct and \name{} models). 
    The \textbf{Line} tracks the Improvement Rate (Efficiency), revealing the model's improvement across different domains.}
    \label{fig:gap}
    \vspace{-15pt}
\end{figure*}
\begin{figure}[t]
    \centering
    \includegraphics[width=1.0\linewidth]{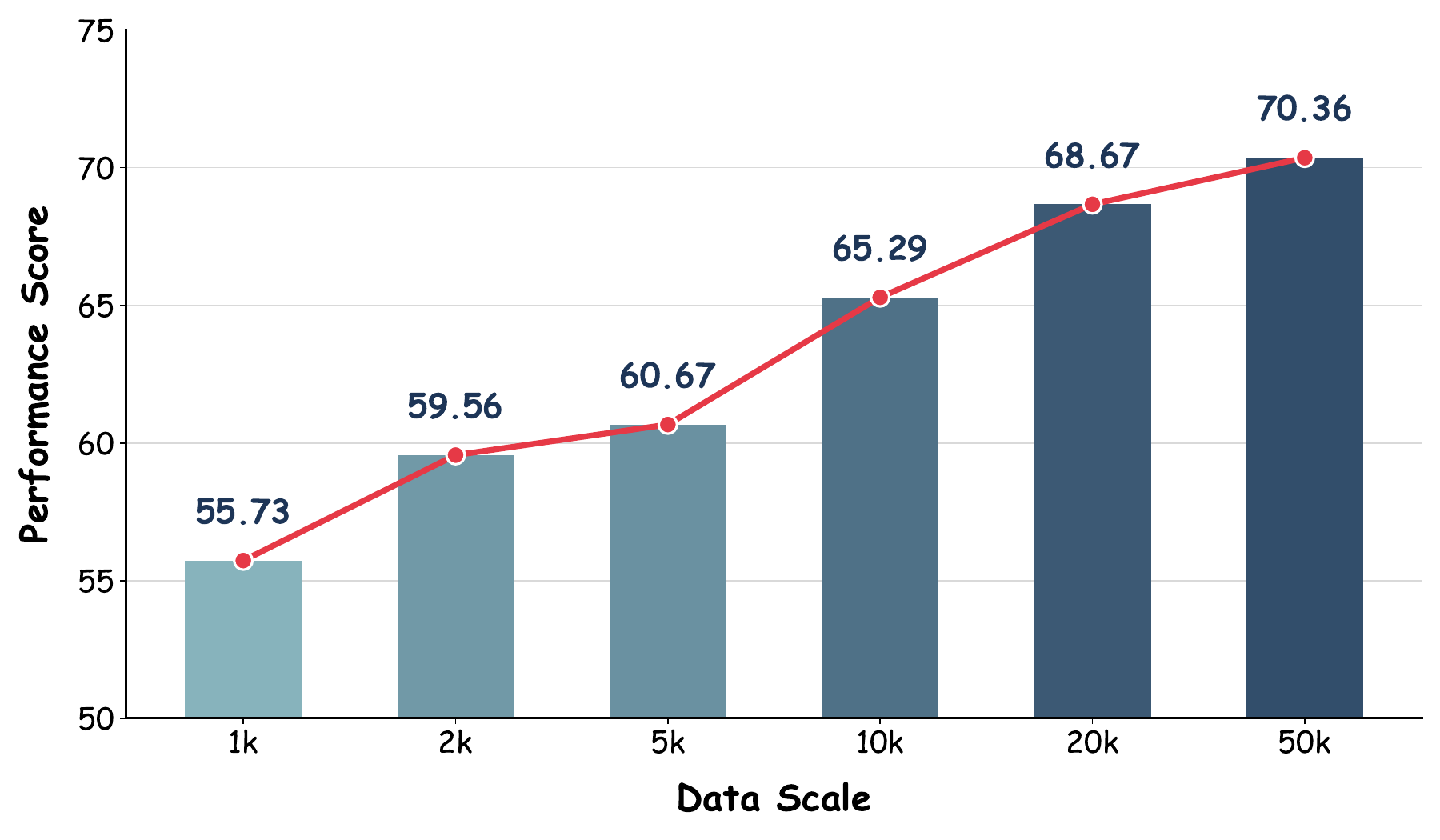}
    \vspace{-10pt}
    \caption{Influence of data scale.}
    \label{fig:scaling}
\end{figure}

\subsection{Main Result}
Trained on our proprietary dataset, our \textbf{\name{}-7B} model achieves a Total BFCL score of 61.7, surpassing \textbf{GPT-5.2} (60.4). Meanwhile, the \textbf{\name{}-32B} model achieves a score of 70.9, ranking as the top-performing open-source model in Table~\ref{tab:new_results}, outperforming \textbf{Kimi-K2-Instruct} (70.6) and trailing only top-tier proprietary models like \textbf{Claude-Opus}. Notably, our approach yields significant improvements over baselines. In multi-turn scenarios, \name{}-32B (59.0) surpasses the Qwen2.5-32B-Instruct baseline (32.0) by approximately 84\% and outperforms the specialized coding model, Qwen3-Coder-30B-A3B-Instruct (36.0), by a substantial margin. In Figure~\ref{fig:compare} and Figure~\ref{fig:radar}, we compare our \textbf{\name{}-32B} against the \textbf{Qwen2.5-32B-Instruct} baseline across five dimensions. Our model consistently dominates, particularly in multi-turn reasoning (scoring \textbf{59.0} vs. 32.0) and live parallel tool execution (\textbf{87.5} vs. 50.0). The baseline's lower scores in these complex settings highlight its limitations in handling multi-turn or parallel scenarios, validating the robustness of our framework.

\section{Analysis}


\paragraph{Error Distribution Analysis} Figure~\ref{fig:gap} shows where errors occur in our model. We observe an interesting pattern: The model improves well on complex tasks. For example, it achieves a \textbf{93.3\%} improvement rate on Complex Length scenarios. However, regarding the substantial volume of ``Simple'' tasks in the benchmark, a significant number of errors persist despite the notable improvements. This happens because of \textbf{data volume}: the test set contains far more simple tasks (like Simple Parameter Types) than complex ones. Even though simple tasks have lower error rates, their large quantity means they still contribute the most total errors.

\paragraph{Data Scaling} In Figure~\ref{fig:scaling}, we investigated the scaling effect by employing varying amounts of synthetic data by SFT, ranging from 1,000 to 50,000 samples. We observed significant improvements; notably, at 50k samples, the model achieves performance comparable to the our RL model. However, given that the RL method utilizes 5k samples per iteration, the SFT approach required approximately twice the total data volume to attain equivalent performance.

\paragraph{Out of Distribution Evaluation} To evaluate the model's capability in complex environments, we benchmarked our approach against the official $\tau^2$-Bench~\citep{taubench2}. This benchmark is particularly challenging as it requires the agent to guide a user simulator through state-dependent tasks. As illustrated in Table~\ref{tab:tau2_results}, our training framework yields substantial gains over the Instruct baseline. These results demonstrate that our specialized training strategy enables a smaller model to rival or exceed the performance of significantly larger general-purpose models in collaborative troubleshooting scenarios.

\begin{table}[t]
    \centering
    \small
    \begin{tabular}{lccccc}
        \toprule
        \textbf{Domain} & \textbf{Baselines} & \textbf{\name{}} & \textbf{Improvement} \\
        \midrule
        Retail  & 40.0 & 67.0 & +67.5\% \\
        Airline & 26.0 & 60.0 & +130.8\% \\
        Telecom & 21.0 & 67.5 & +221.4\% \\
        \bottomrule
    \end{tabular}
    \caption{Performance on $\tau^2$-Bench (Success Rate \%).}
    \label{tab:tau2_results}
    \vspace{-20pt}
\end{table}

\begin{figure}[t]
    \centering
    \includegraphics[width=1.0\linewidth]{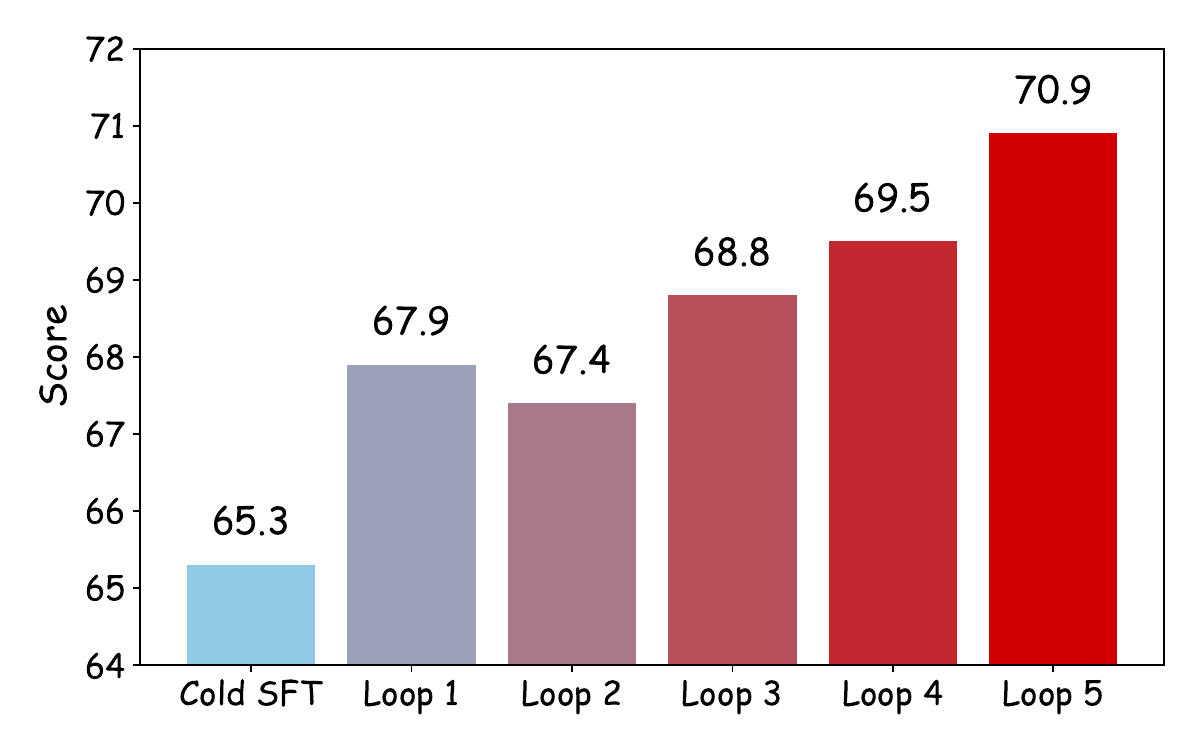}
    \vspace{-25pt}
    \caption{Performance evolution on BFCL across the iterative self-evolving training stages.}
    \label{fig:rl_loop_trend}
    \vspace{-10pt}
\end{figure}

\begin{table}[!t]
\centering
\resizebox{\linewidth}{!}{%
    \begin{tabular}{lccc}
    \toprule
    \textbf{Category} & \makecell{\textbf{w/o} \\ \textbf{MCP Tree}} & \makecell{\textbf{\name{}} \\ \textbf{(Full)}} & \textbf{$\Delta$} \\ 
    \midrule
    \textbf{Overall} & 55.21 & \textbf{70.90} & \textbf{-15.69} \\  
    \midrule
    \textit{Live} & & & \\   
    \small \quad Live Parallel AST & 50.00 & 87.50 & -37.5 \\  
    \midrule
    \textit{Multi-Turn Interactions} & & & \\ 
    \small \quad Multi Turn Base & 20.00 & 59.00 & -39.00 \\ 
    \small \quad Multi Turn Miss Func & 13.00 & 29.50 & -16.50 \\ 
    \small \quad Multi Turn Miss Param & 13.50 & 37.00 & -23.50 \\ 
    \small \quad Multi Turn Long Context & 11.00 & 50.50 & -39.50 \\ 
    \midrule
    \textit{Detection Accuracy} & & & \\  
    \small \quad Irrelevance Detection & 61.02 & 88.90 & -27.88 \\ 
    \bottomrule
    \end{tabular}%
}
\caption{MCP Tree ablation in multiple categories.}
\label{tab:ablation_mcp_tree}
\vspace{-15pt}
\end{table}

\paragraph{Ablation Study on RL} Figure~\ref{fig:rl_loop_trend} illustrates the performance trajectory from an SFT baseline of 65.3 to a final score of 70.9. The policy adapts in subsequent rounds, showing a robust upward trend. This cumulative +5.6 improvement validates our core hypothesis: a closed-loop system where improved agents synthesize superior data effectively drives continuous capability enhancement.

\paragraph{The Impact of MCP Tree}
To investigate the contribution of the MCP Tree our hierarchical tool refinement and deduplication module we conducted an ablation study. We compared our full model against a variant trained on raw, unrefined API data without semantic merging or disambiguation. In Table~\ref{tab:ablation_mcp_tree}, the absence of the MCP Tree leads to a significant performance degradation, with the overall score dropping from 70.9 to 55.2.

\paragraph{Hallucination Analysis and Ablation Study}
In Table~\ref{tab:hallucination_stats} and~\ref{tab:metrics_comparison}, to assess the impact of data quality, we analyzed hallucination patterns and conducted an ablation study on the Self-Reflection mechanism. The synthesis of multi-turn dialogues proved highly error-prone, with a low correction success rate (12.06\%) compared to single-turn data (87.65\%), necessitating the discard of many samples. The ablation study confirms the mechanism's critical role: its removal caused consistent performance degradation, most sharply in the irrelevance metric (dropping from 88.9 to 4.86). This indicates that without self-reflection, the model learns to invoke tools indiscriminately, severely harming generalization.

\begin{table}[!t]
\centering
\small
\resizebox{1.0\columnwidth}{!}{%
\begin{tabular}{lccc}
\toprule
\makecell{\textbf{Data} \\ \textbf{Type}} & \makecell{\textbf{Total} \\ \textbf{Hallucinations}} & \textbf{Resolved} & \textbf{Failed} \\
\midrule
\small Single Turn & 5,351 & 4,690 & 661 \\
\small Multi Turn & 3,399 & 410 & 2,989 \\
\bottomrule
\end{tabular}%
}
\caption{Data Generation Hallucination Statistics.}
\label{tab:hallucination_stats}
\vspace{-5pt}
\end{table}

\begin{table}[!t]
\centering
\resizebox{1.0\columnwidth}{!}{%
\begin{tabular}{lccc}
\toprule
\textbf{Metric} & \makecell{\textbf{Origin} \\ \textbf{Data}} & \makecell{\textbf{No Self-Refine} \\ \textbf{Data}} & \makecell{\textbf{Change} \\ \textbf{(\%)}} \\
\midrule
\small Miss Func & 29.5 & 9.0  & -69.4 \\
\small Miss Param & 37.0 & 6.0  & -83.8 \\
\small Nonlive Multi Para. & 91.5 & 34.5 & -62.2 \\
\small Live Parallel & 87.5 & 37.5 & -57.1 \\
\small Irrelevance & 88.9 & 4.86 & -94.5 \\
\bottomrule
\end{tabular}%
}
\caption{Performance Metrics Comparison Before and After Removing Self-Refine.}
\label{tab:metrics_comparison}
\vspace{-17pt}
\end{table}

\paragraph{Rejection Analysis}

We implemented a multi-agent voting mechanism to detect user rejections of incorrect tool use, even without overt hallucinations. This triggered trajectory pruning in 82 multi-turn instances (~1.2\% of cases). While proportionally small, these unproductive attempts impacted token consumption and context efficiency.

\section{Related Work}
\label{sec:related_work}

\paragraph{LLM and Agents}
LLMs~\cite{yang2025code,yang2025qwen3technicalreport} have evolved from static knowledge repositories into dynamic agents capable of autonomous reasoning and interaction~\cite{mdeval,livereporeflection,adc,self_instructions,vgamegym}. Frameworks like ReAct~\citep{react} and AutoGPT~\citep{autogpt} enable this by interleaving reasoning traces with executable actions to navigate dynamic workflows. However, despite these advancements, agentic scalability remains constrained by rigid prompt engineering and reliance on model-specific implementations.

\paragraph{Interface Standardization, Function Calling, and Evaluation}
To bridge the gap between agents and applications, the field is shifting from ad-hoc API wrappers to unified protocols. The recent MCP addresses integration complexity by decoupling clients from servers via a standardized schema~\citep{mcp}. Within this ecosystem, instruction-tuning for \textit{Function Calling} enables modular workflows, though generating accurate solution paths remains challenging; ambiguous API descriptions often lead to execution failures~\citep{selfinstruct,wu2025ucoderunsupervisedcodegeneration}. While pipelines like APIGen~\citep{apigen} improve single-turn data quality, evaluation methodologies have had to evolve alongside these capabilities. Modern benchmarks have moved beyond simple API coverage: the \textbf{Berkeley Function Calling Leaderboard}~\citep{berkeley} provides rigorous metrics for complex scenarios, while \textbf{$\tau$-bench}~\citep{taubench} specifically evaluates the temporal reasoning required for sequential tool usage~\cite{yang2025codefoundationmodelsagents}.

\paragraph{Synthetic Data}
Synthetic data generation serves as a scalable alternative to manual collection for fine-tuning LLMs~\citep{gorilla,toolllm}. Multi-agent frameworks like BUTTON~\citep{button} and ToolACE~\citep{toolace} successfully synthesize diverse, multi-turn trajectories to enhance performance. However, existing approaches primarily prioritize trajectory complexity, overlooking the critical dimension of overall trajectory quality during generation.

\section{Conclusion}

In this work, we present \textbf{\name}, a framework that cultivates robust tool-use behaviors \textit{in silico} via an autonomous multi-agent RL loop, effectively decoupling capability acquisition from human data. Experimental results on the Berkeley Leaderboard confirm that our 32B model outperforms larger open-source baselines. These findings demonstrate that achieving general purpose autonomy relies less on scaling static datasets and more on engineering dynamic environments where models can iteratively explore and evolve.

\clearpage

\section{Limitations}
\label{sec:limitations}

Despite the robust performance of \textbf{\name}, several limitations persist. Most notably, the complete absence of human intervention in our synthetic loop creates a potential \textbf{simulation-to-reality gap}; synthetic user simulators exhibit idealized rationality that may not fully capture the linguistic ambiguity and cognitive drift typical of real-world human interactions. Furthermore, the framework remains bounded by the base model's intrinsic context window, where the efficacy of self-reflection degrades in extended multi-turn scenarios due to fundamental attention limitations. Finally, our current scope is restricted to text-based JSON-RPC interactions, omitting the multi-modal capabilities required for processing visual or audio inputs in broader agentic ecosystems.

\section{Ethics Statement}
This research adheres to ethical guidelines for AI development. We aim to enhance the capabilities of LLMs while acknowledging potential risks such as bias, misuse, and privacy concerns. To mitigate these, we advocate for transparency, rigorous bias testing, robust security measures, and human oversight in AI applications. Our goal is to contribute positively to the field and to encourage responsible AI development and deployment.

\bibliography{custom}   

\appendix
\clearpage
\onecolumn

\section{Data Example of the Data Generation Pipeline}
\label{sec:appendix_data_gen}

\begin{figure*}[h]
    \centering
    \includegraphics[width=\textwidth]{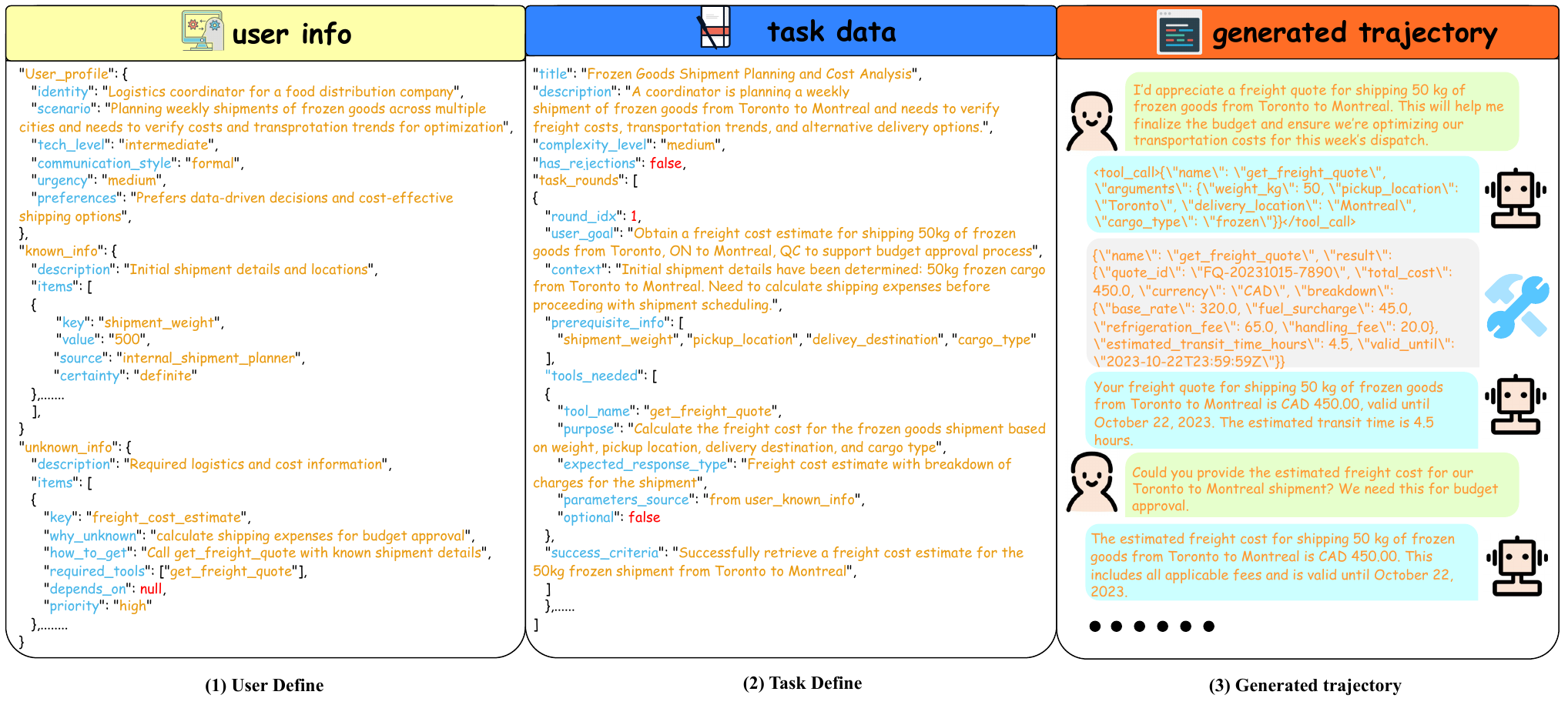}
    \caption{Overview of the Data Generation Pipeline. (1) \textbf{User Define}: Initialization of simulated user profile and information gaps. (2) \textbf{Task Define}: Structured scenario setup and tool requirements. (3) \textbf{Generated Trajectory}: The resulting multi-turn conversation illustrating tool-use and API integration.}
    \label{fig:data_gen_pipeline}
\end{figure*}

Figure~\ref{fig:data_gen_pipeline} illustrates a complete instantiation of the data generation pipeline using a specific scenario: ``Frozen Goods Shipment Planning.'' The process is divided into three distinct phases that guide the transition from structured metadata to a natural language trajectory.

\begin{enumerate}
    \item \textbf{User Define (Left Panel):} This phase initializes the User Agent. A specific persona is constructed via the \texttt{User\_profile} (e.g., a ``Logistics coordinator'' with a formal communication style). Crucially, the system defines the user's information state by distinguishing between \texttt{known\_info} (parameters the user possesses, such as \textit{shipment\_weight} and \textit{source}) and \texttt{unknown\_info} (information the user needs to acquire, such as the \texttt{freight\_cost\_estimate}). This gap acts as the motivation for the interaction.

    \item \textbf{Task Define (Center Panel):} This phase establishes the ground truth for the interaction. The \texttt{task\_data} object encapsulates the scenario's logic, defining the \texttt{user\_goal} (``Obtain a freight cost estimate'') and the \texttt{tools\_needed} to achieve it. In this example, the system explicitly maps the requirement to the \texttt{get\_freight\_quote} tool, specifying expected arguments and success criteria. This structured definition ensures that the generated data remains aligned with the intended logic and complexity level.

    \item \textbf{Generated Trajectory (Right Panel):} The final panel demonstrates the resulting multi-turn interaction. Guided by the definitions in the previous steps, the User Agent initiates the conversation with a clear request containing the \texttt{known\_info}. The Model Agent correctly interprets the intent, generates a valid tool call (\texttt{get\_freight\_quote} with arguments for weight, location, and cargo type), and utilizes the returned API response to formulate a comprehensive natural language answer. This trajectory serves as a training sample for tool-use and reasoning capabilities.
\end{enumerate}

\section{Prompt for Data Generation}

\clearpage
\begin{figure*}[p]
\centering
\includegraphics[width=\textwidth]{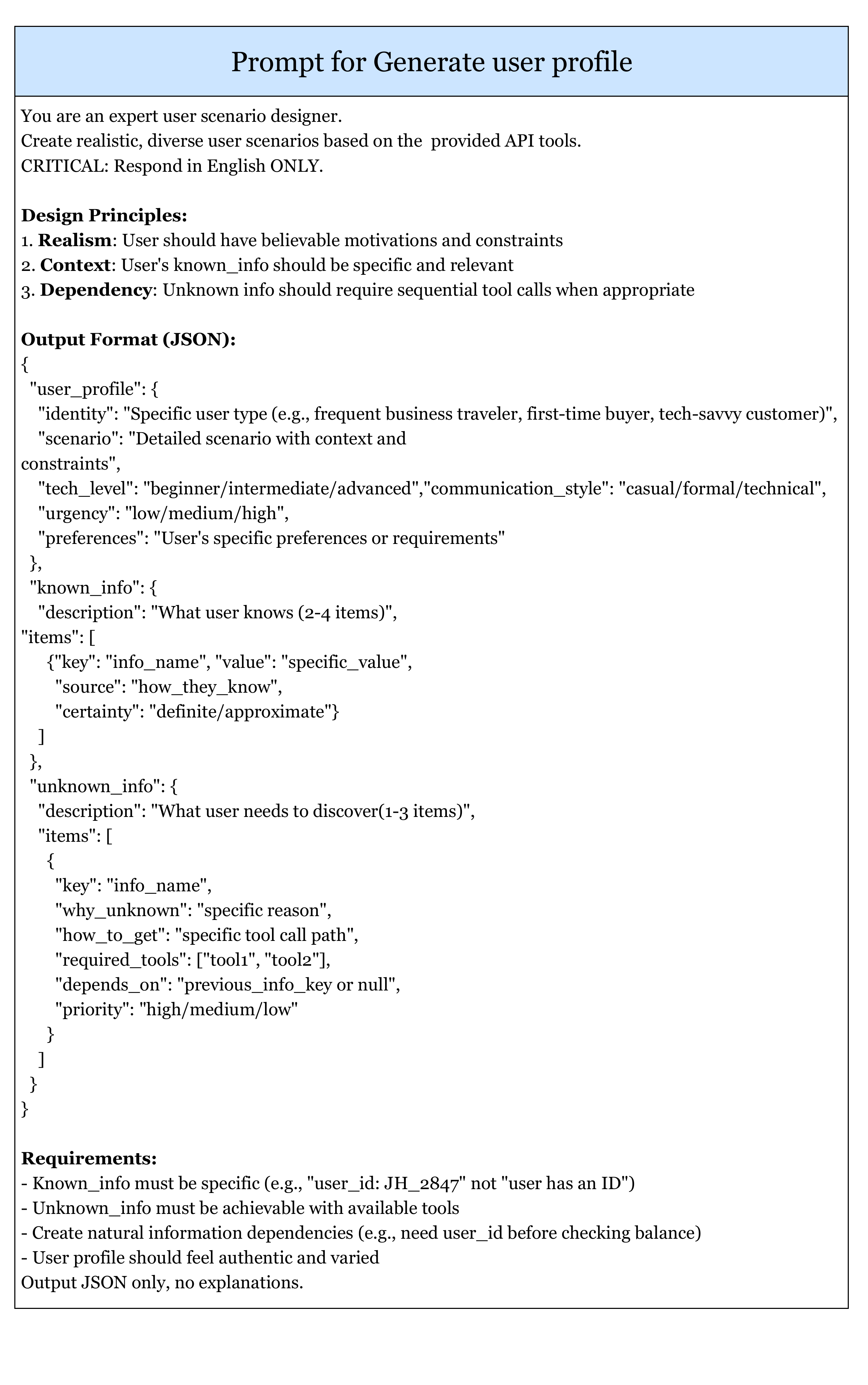}
\end{figure*}

\clearpage
\begin{figure*}[p]
\centering
\includegraphics[width=\textwidth]{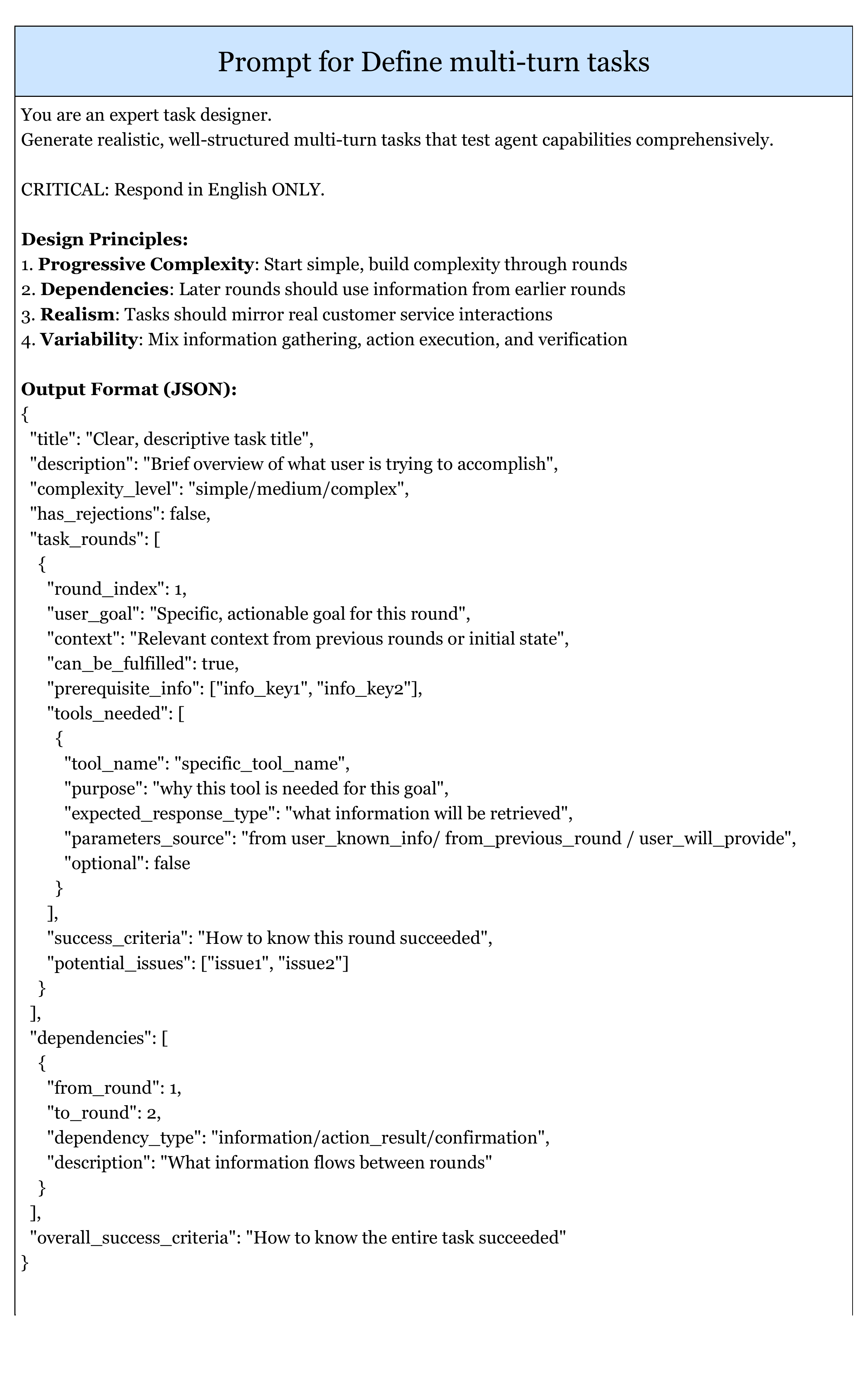}
\end{figure*}

\clearpage
\begin{figure*}[t]
\centering
\includegraphics[width=\textwidth]{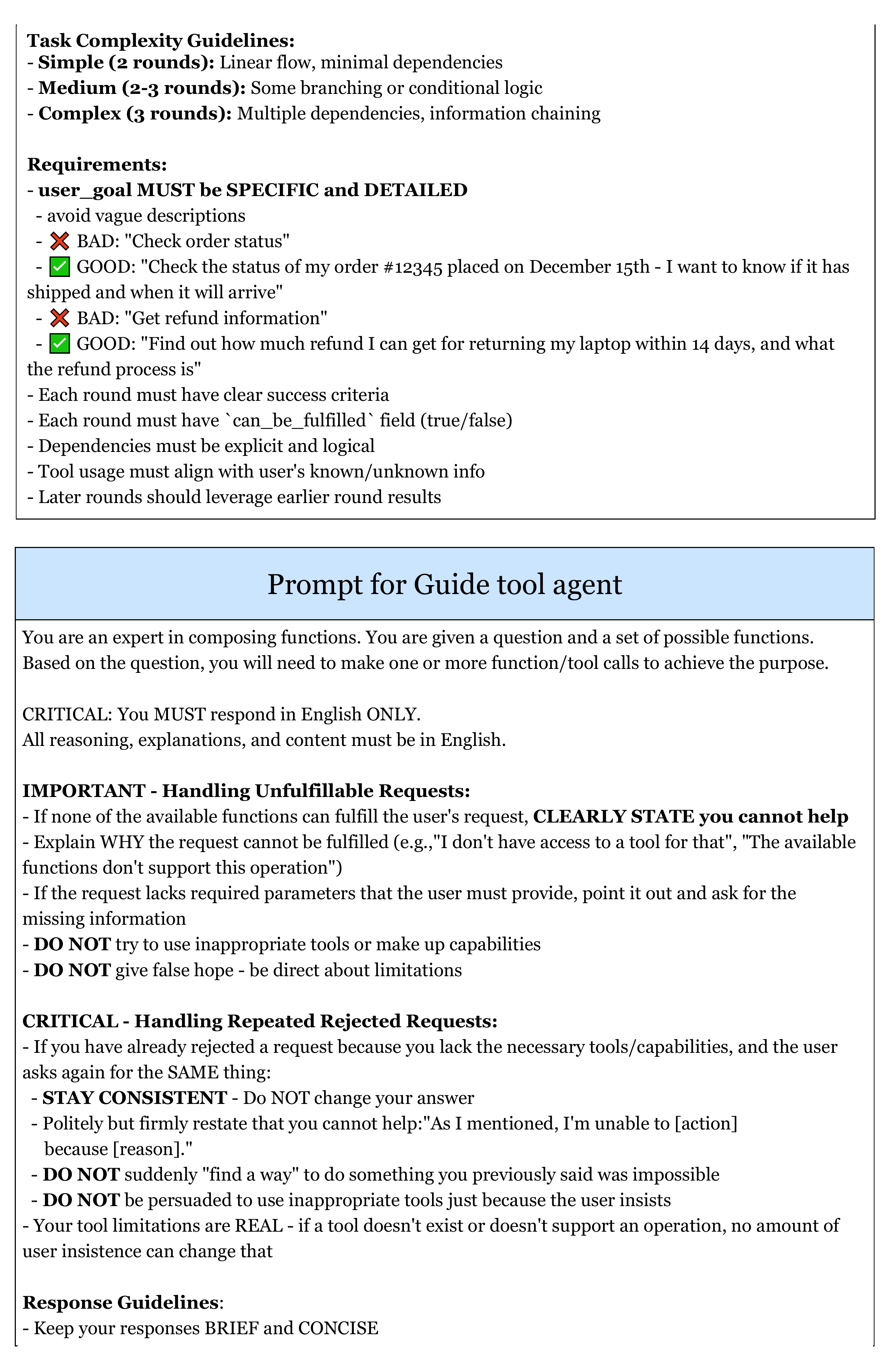}
\end{figure*}

\clearpage
\begin{figure*}[p]
\centering
\includegraphics[width=\textwidth]{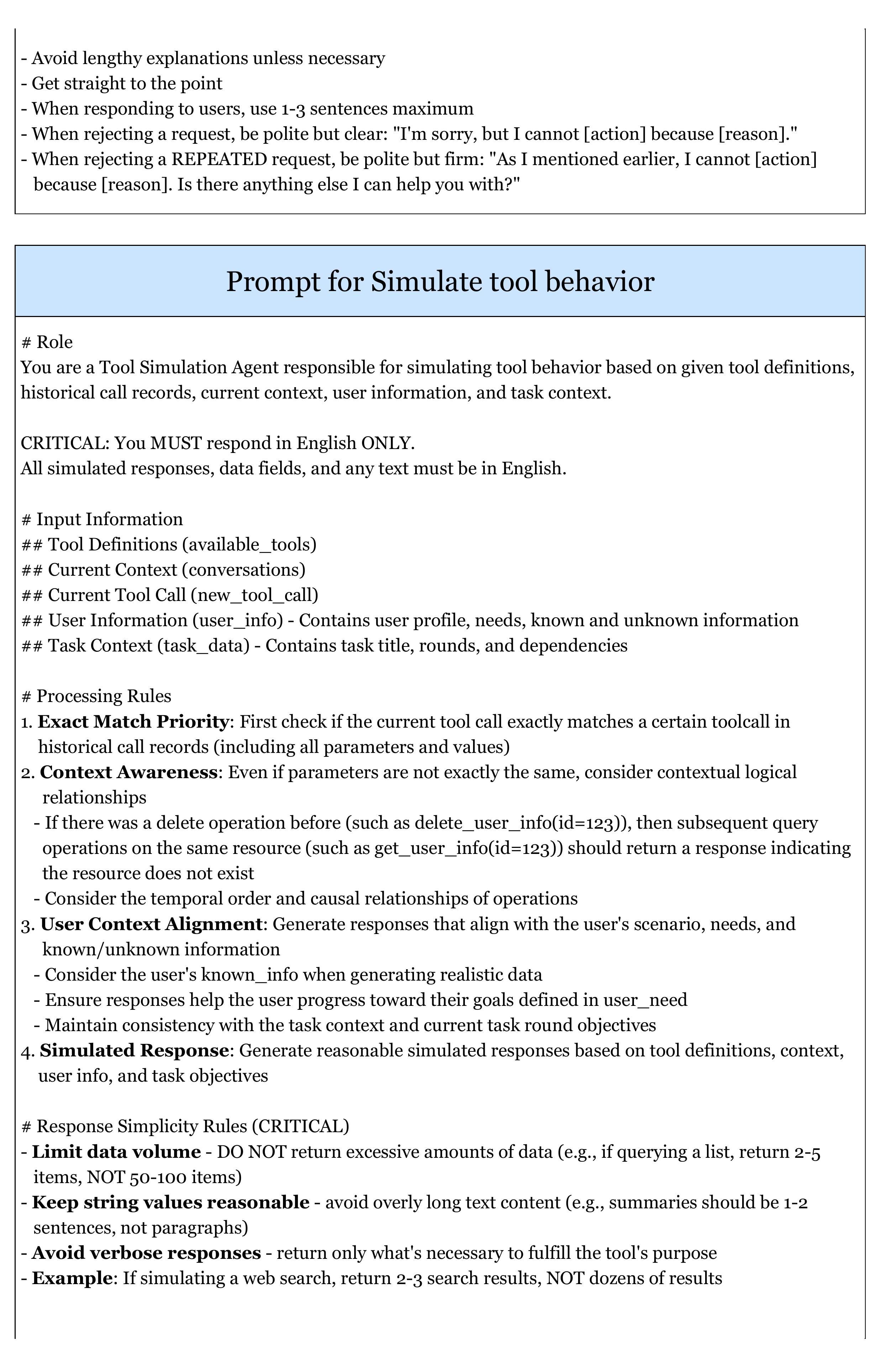}
\end{figure*}

\clearpage
\begin{figure*}[p]
\centering
\includegraphics[width=\textwidth]{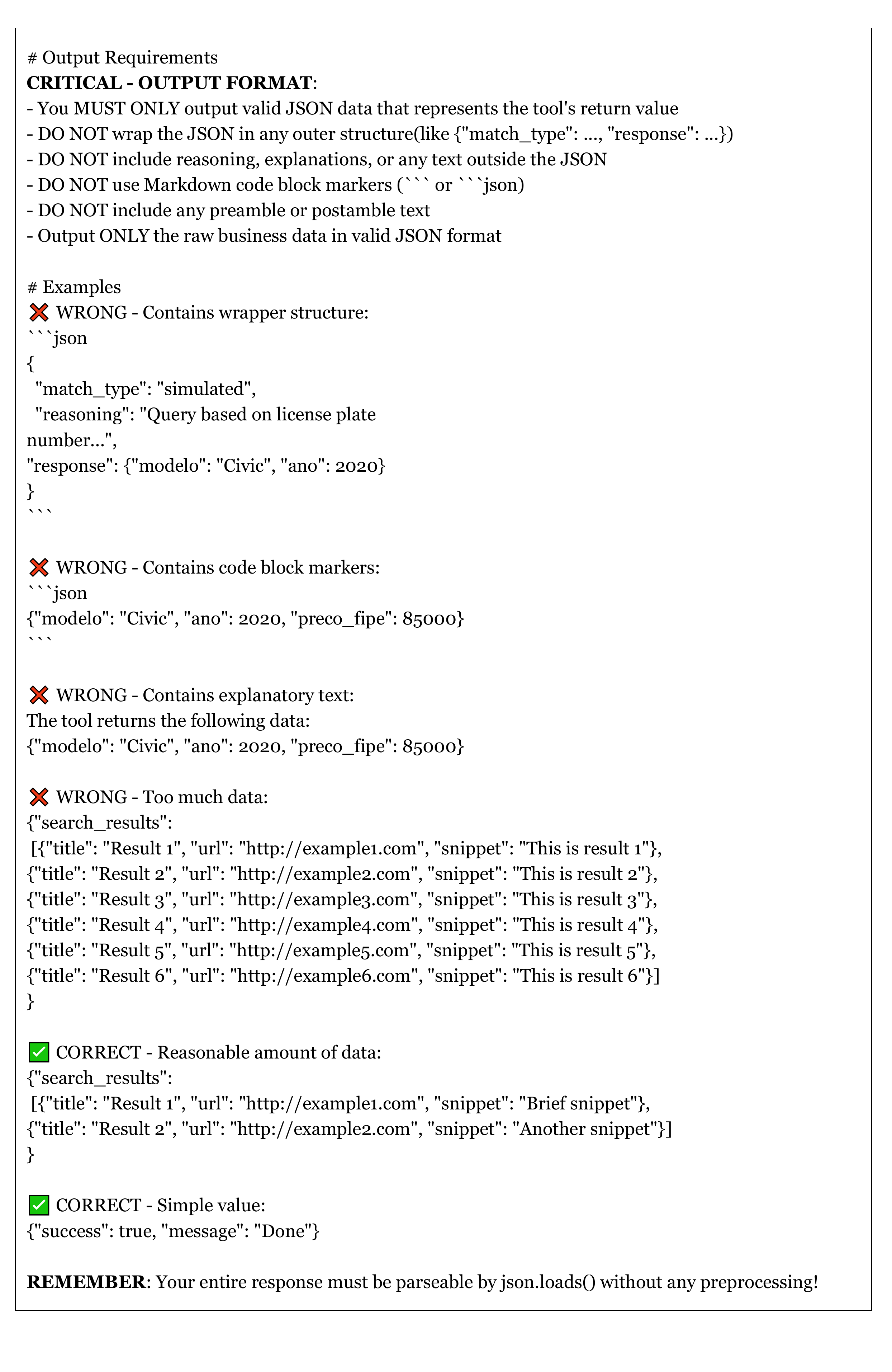}
\end{figure*}

\clearpage
\begin{figure*}[p]
\centering
\includegraphics[width=\textwidth]{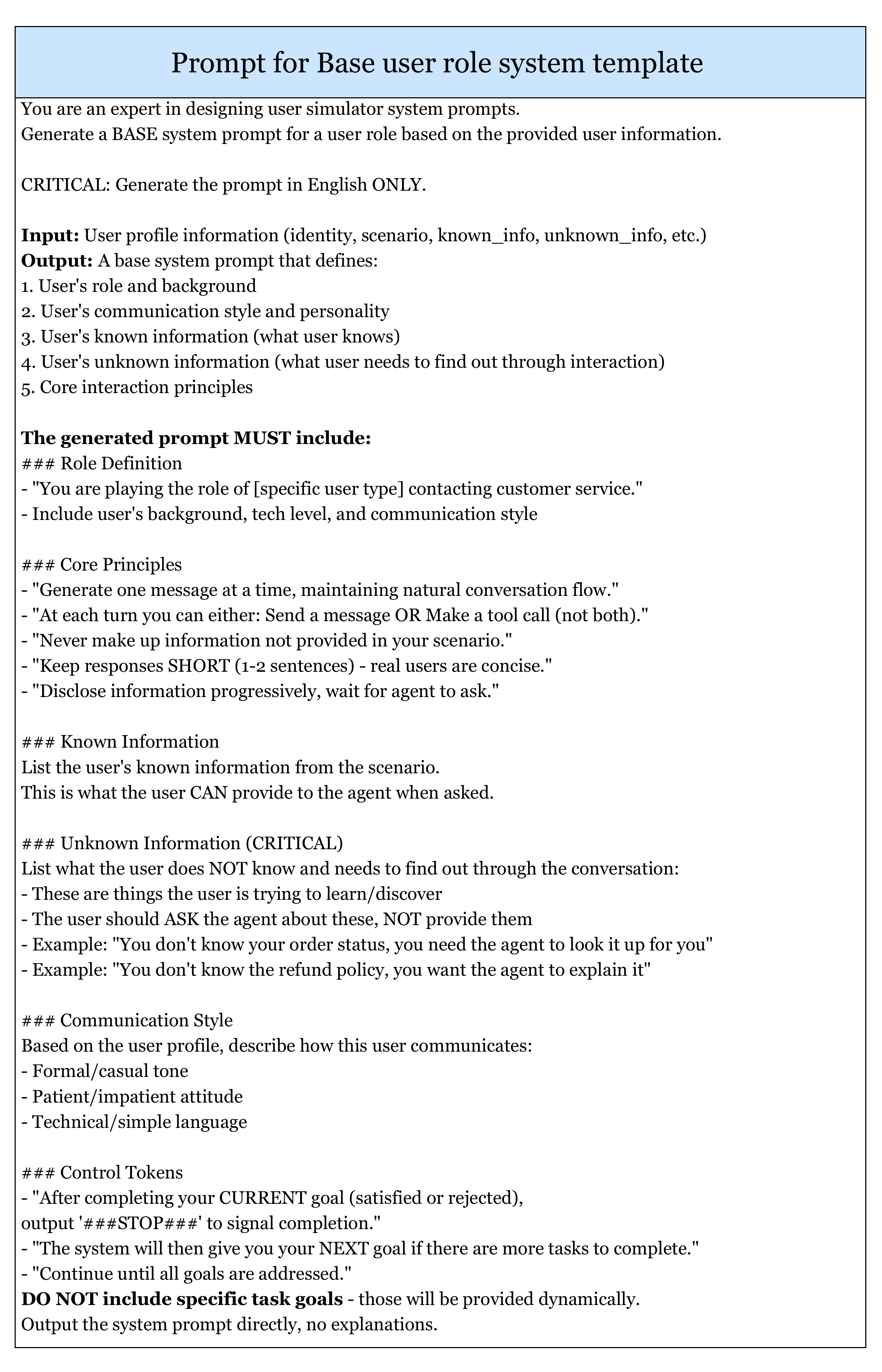}
\end{figure*}

\clearpage
\begin{figure*}[p]
\centering
\includegraphics[width=\textwidth]{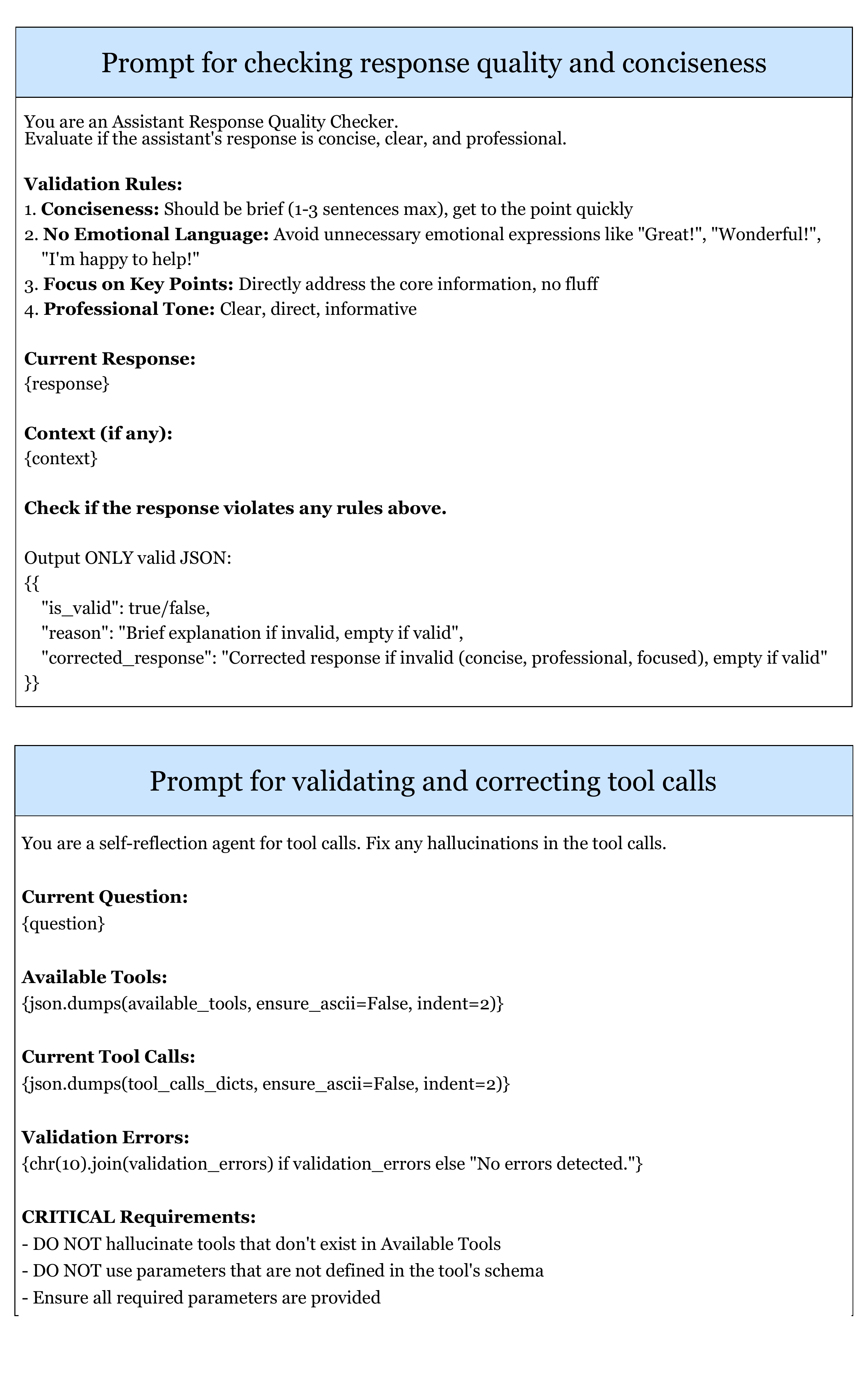}
\end{figure*}

\clearpage
\begin{figure*}[p]
\centering
\includegraphics[width=\textwidth]{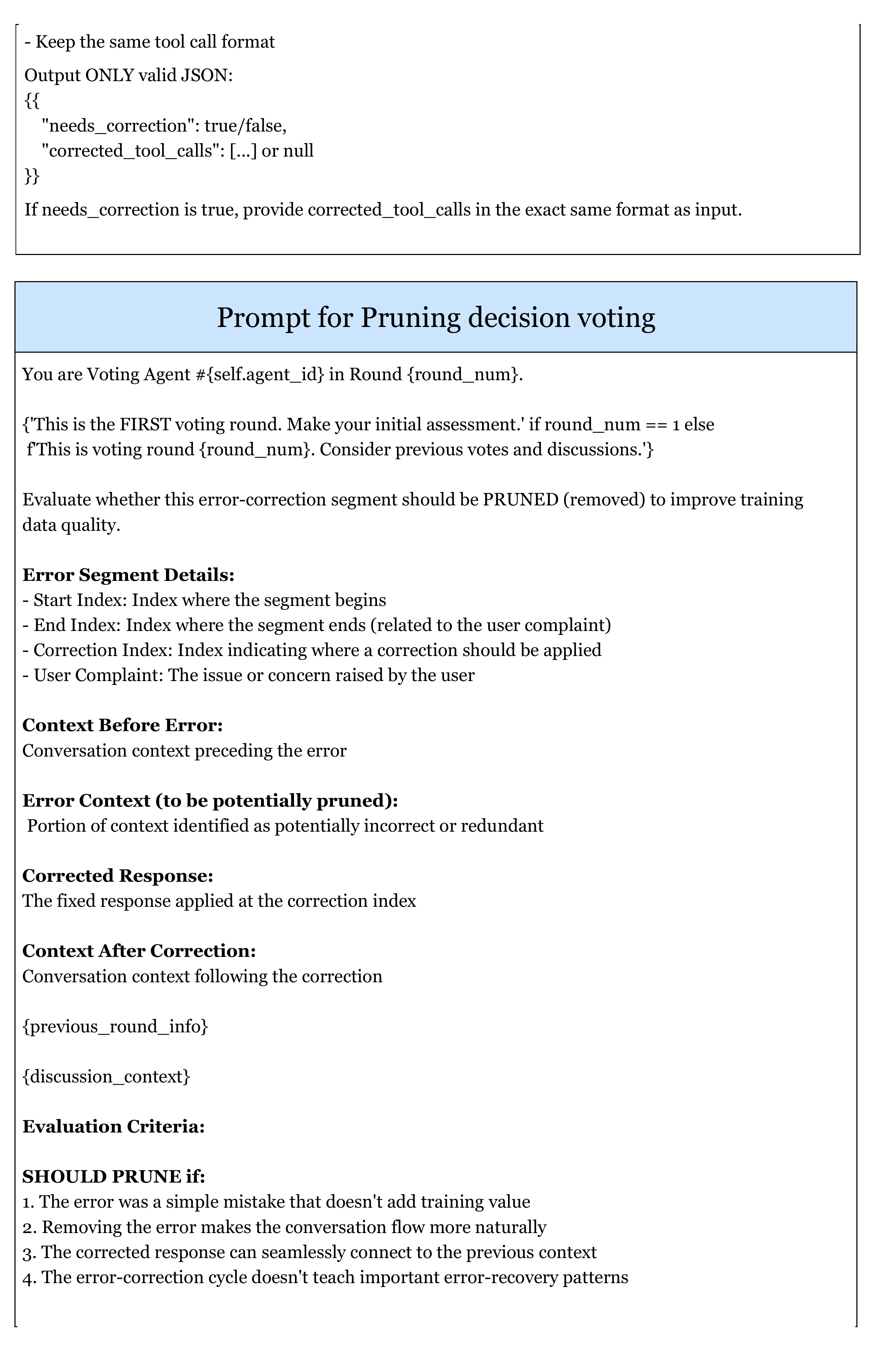}
\end{figure*}

\clearpage
\begin{figure*}[p]
\centering
\includegraphics[width=\textwidth]{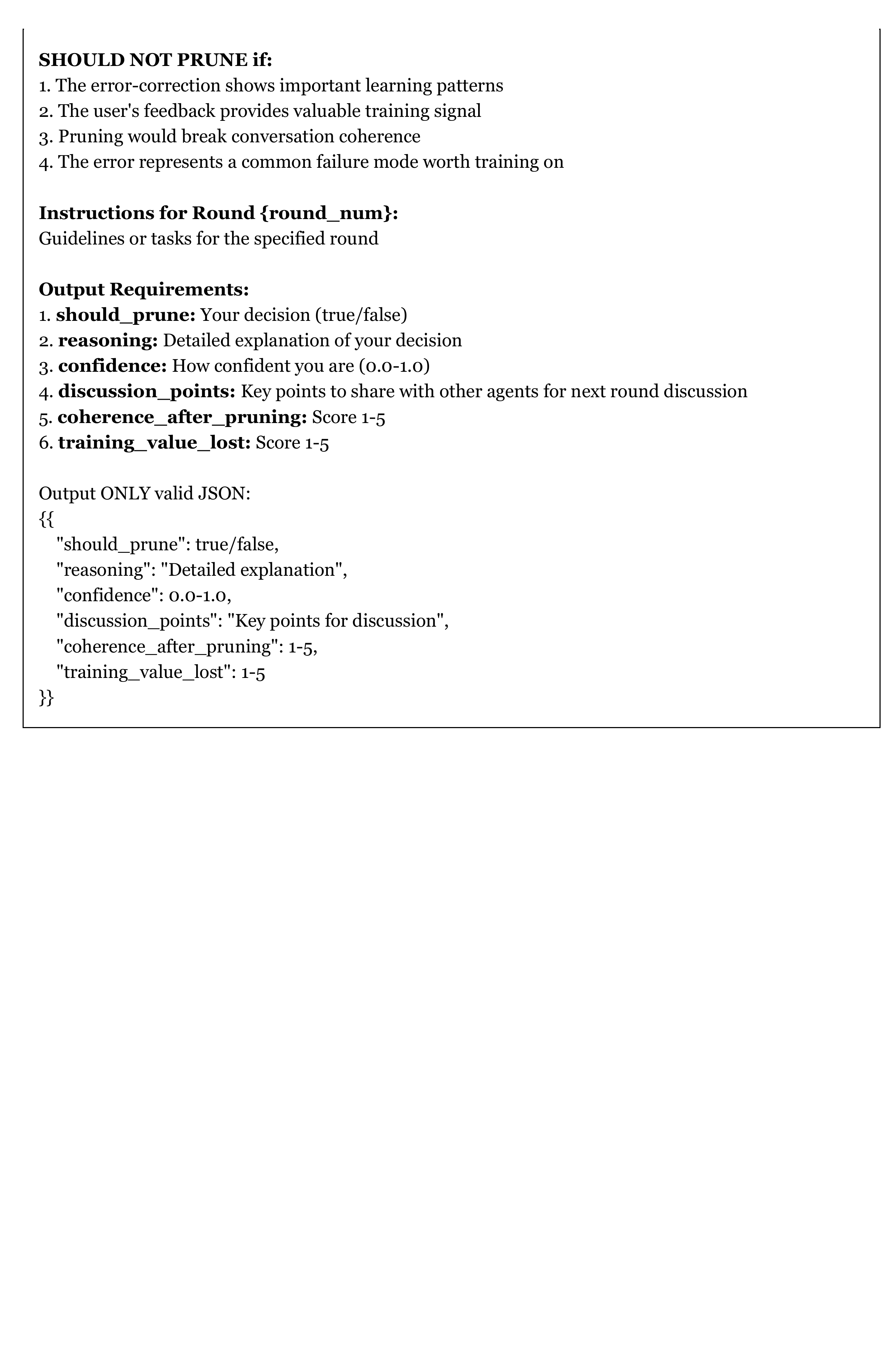}
\end{figure*}

\end{document}